\documentclass[onecolumn]{article}         
\usepackage{etex}
\usepackage[top=3.6cm, bottom=3.2cm, left=2.5cm, right=2.5cm]{geometry}

\usepackage{mathrsfs}
\usepackage{algorithm}
\usepackage{algpseudocode}
\usepackage{bm}
\usepackage{graphicx}
\usepackage{subfigure}
\usepackage{amssymb}
\usepackage{amsmath}
\usepackage[table]{xcolor}
\usepackage{array,color}
\usepackage{arydshln}
\usepackage{verbatim}
\usepackage{booktabs}
\usepackage{tikz}
\usepackage{ctable}
\usepackage{multirow}
\usepackage{multicol}
\usepackage{rotating}
\usepackage{threeparttable}
\usepackage{cite}
\usepackage{pbox}
\usepackage[colorlinks=true,linkcolor=red,citecolor=blue]{hyperref}
\usepackage{url}

\newcommand*{\DashedArrow}[1][]{\mathbin{\tikz [baseline=-0.25ex,-latex, dashed,#1] \draw [#1] (0pt,0.5ex) -- (1.3em,0.5ex);}}%

\newtheorem{myDef}{Definition}

\begin{document}
%
\title{Texture Characterization by Using Shape Co-occurrence Patterns}

\author{Gui-Song~Xia$^1$, Gang Liu$^2$, Xiang~Bai$^3$, and Liangpei~Zhang$^1$\\
$^1${\em State Key Lab. LIESMARS, Wuhan University, Wuhan, China.}\\
$^2${\em CNRS LTCI, Telecom ParisTech, Paris, France.}\\
$^2${\em Electronic Information School, Huazhong University of Science and Technology, China.}
}

\maketitle

\begin{abstract}
Texture characterization is a key problem in image understanding and pattern recognition. In this paper, we present a flexible shape-based texture representation using shape co-occurrence patterns. More precisely,  texture images are first represented by tree of shapes, each of which is associated with several geometrical and radiometric attributes. Then four typical kinds of shape co-occurrence patterns based on the hierarchical relationship of the shapes in the tree are learned as codewords. Three different coding methods are investigated to learn the codewords, with which, any given texture image can be encoded into a descriptive vector. In contrast with existing works, the proposed method not only inherits the strong ability to depict geometrical aspects of textures and the high robustness to variations of imaging conditions from the shape-based method, but also provides a flexible way to consider shape relationships and to compute high-order statistics on the tree. To our knowledge, this is the first time to use co-occurrence patterns of explicit shapes as a tool for texture analysis. Experiments on various texture datasets and scene datasets demonstrate the efficiency of the proposed method.

\end{abstract}


\section{Introduction}
As a fundamental ingredient of image structures, texture conveys important cues in numerous processes of human visual perception. While, due to the high complexity of the structures in natural images, the modeling of texture is a challenging problem in image analysis and understanding.

This paper addresses the problem of texture characterization, one difficulty of which lies in the fact that: ``\emph{texture is a stuff that is easy to recognize but difficult to define}"~\cite{Jeanponce02computervision}.
Over the years, tremendous investigations have been made on texture analysis, see e.g.~\cite{texture_Haralick73,LeungM01,Portilla2000,texture_Lazebnik2,xia2010shape,Pedersen2013,Li2013,Matthews2013,sifre2013rotation,liu2011sorted,liu2012, xu2010new}, based on the main observation from texture perception which suggested that human texture discrimination could be modeled by statistics of ``a few local conspicuous features", so-called \emph{textons}~\cite{Julesz}.
Among them three aspects of textures have been mainly addressed :
\begin{itemize}
\item[-] \emph{{How to depict the geometrical aspects of textures}}.
Based on the experiments of texture perception, Julsez~\cite{Julesz} suggested that human texture discrimination could be modeled by statistics of  \emph{textons}, such as edges, line ends, blobs, etc. Several discriminative textons features found by Julesz include the closure, connectivity and granularity of the local geometries of images.
In order to represent the structural aspects of textures, some mathematical tools, such as Gabor or wavelet-like analysis, are used to probe the atomic texture elements such as elongated blobs and terminators in images, and the marginal/joint distributions of the resulted responses are subsequently utilized to describe the statistical arrangement of texture~\cite{LeungM01,Portilla2000,Zhu2005textons}. The strong ability of such mathematical tools to handle multi-scale and oriented structures has made them one of the most popular tool for texture analysis.
However, how to efficiently represent the highly geometrical aspects of textures, e.g. sharp transitions and elongated contours, is an open issue. To solve this problem, alternative wavelet-like approaches, e.g. Grouplet~\cite{Peyre08} and scattering transform~\cite{sifre2013rotation}, have been elaborated to enable more efficient representations of structured textures.
In contrast with explicit models, patch-based method~\cite{Varma2005} provides another possibility for describing the structured aspects of textures, but it is not trivial to capture the multi-scale nature of textures by patch-based method.

\item[-] \emph{How to capture the invariant properties of textures}
Next, for meeting the requirement of invariance with respect to viewpoint and illumination changes,
many invariant texture descriptors have been proposed in the literature, including the rotation invariant local binary pattern (joint distribution of gray values on circular local neighborhoods)~\cite{Ojala2002}, the multi-fractal analysis method~\cite{xu2009viewpoint, xu2010new,ji2013wavelet}, \emph{etc.}.
Recently, several approaches rely on the extraction of local features that are individually invariant to some geometric transforms, such as scaling, rotating and shearing~\cite{texture_Lazebnik2}. Compared with previous works dealing with invariant texture analysis, such locally invariant methods do not need any learning of the deformations.
Alternatively, by relying on morphological operations, the shape-based invariant texture analysis method~\cite{xia2010shape}, named by SITA for short in this paper, represents a texture by a tree of explicit shapes and shape attributes are locally normalized to achieve invariant texture description.
It has reported that SITA achieved superior performance on invariant texture recognition. One limitation of this work, however, lies in its difficult to take into account high-order statistics of shapes, which has been demonstrated as a crucial factor for texture discrimination~\cite{xia2010shape}.

\item[-] \emph{How to learn high-order statistics for texture analysis}
{Last but not least, the high-order information can be learned to describe textures from different perspectives, which highly improve the ability to recognize textures. Among these methods, a multivariate log-Gaussian Cox process is applied to model the relationship of key points~\cite{nguyen2011visual}. The pairwise local binary pattern (LBP) are developed in ~\cite{qi2014Pairwise} to depict the relationship between LBP. Besides, the scattering transform net~\cite{sifre2013rotation} and the convolutional neural networks~\cite{Cimpoi2015} use cascaded networks to learn the high-order informations of images for texture characterization. Similar to these methods, the objective of this paper is to present a more flexible shape-based texture analysis framework by investigating the co-occurrence patterns of shapes, which can model the high-order geometrical aspects of textures.
	}

\end{itemize}



\subsection{Related Work}
\label{SEC:rw}
Over the past decades, tremendous investigations have been made in texture analysis, see e.g.~\cite{LeungM01,Portilla2000,texture_Lazebnik2,xia2010shape,Pedersen2013,Li2013,Matthews2013,sifre2013rotation}, among which an active topic is developing texture models which can efficiently depict both the statistical and the geometrical aspects of textures and are robust to the variations of imaging condition as well.

The proposed texture analysis method relying on shape co-occurrence patterns is inspired both by the shape-based texture analysis scheme and texture models using co-occurrence matrix, and is also closely related to the texton-based texture modeling paradigm. In what follows, we briefly recall these backgrounds and related works.

\begin{figure}[ht!]
\centering
  \includegraphics[width=0.5\linewidth]{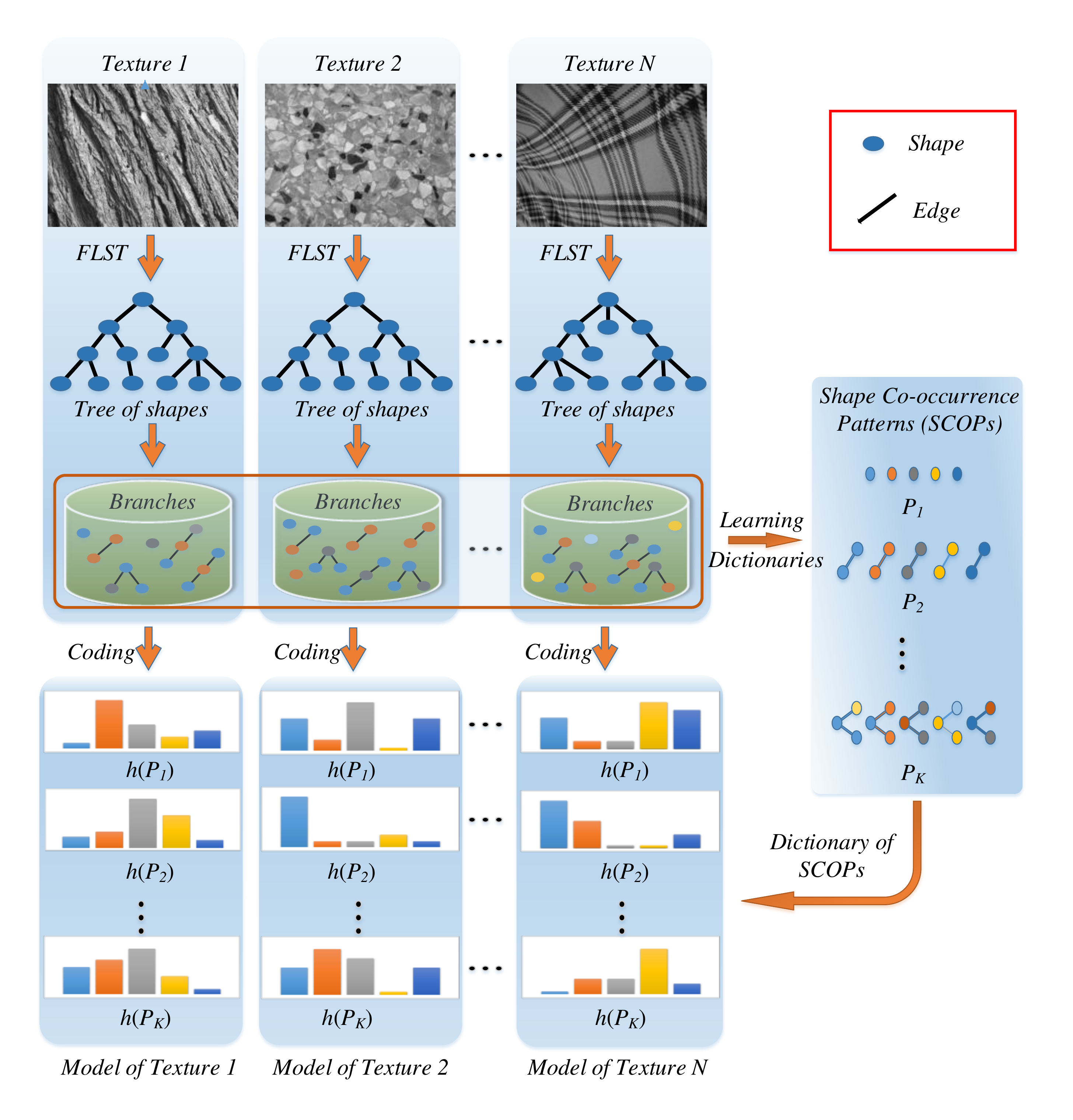}\\
  \caption{The flowchart of our algorithm. In our algorithm, images are represented by tree of shapes(topographic map) via Fast Level Set Transform (FLST) firstly. Then the branches of the tree are collected with a statistical method to learn the shape co-occurrence patterns. Finally, the corresponding histogram feature of the images are extracted by these patterns.}
  \label{fig:flowchart}
\end{figure}

\subsubsection{Texton-based texture analysis paradigm}
Julesz~\cite{Julesz} found that the first-order statistics of ``a few local conspicuous features", called \emph{texton}, are significant for texture discrimination. The texton theory in fact led to a kind of structural approaches to texture
description, which first probes texture primitives as local features and then investigates their organization.
An implicit way to implement the texton theory is to use wavelet-like analysis, such as Gabor filter banks, to probe atomic texture elements in texture images and then utilize the resulted filtering responses to describe the underlying statistical texture features~\cite{LeungM01,Portilla2000,sifre2013rotation}.

An alterative implementation is to explicitly detect atomic texture elements beforehand and model the spatial arrangements of such textons. For instance, Zhu \emph{et al.}~\cite{Zhu2005textons} detected textons by using a number of image bases with deformable spatial configurations, which are learned from static texture images. Lafarge \emph{et al.}~\cite{Lafarge09} first defined a set of geometric objects, e.g. segment, line, circle, band etc., as texture elements and then detected those elements and studied their organization through Markov Point Process (MPP).
Note that a more general manner is to extract small image patches from texture, cluster them into textons and finally investigate the underlying statistics. Lazebnik \emph{et al.}~\cite{lazebnik2005sparse} extended this idea by removing the redundancy between patch-based textons through utilizing interest regions in images.
Compared with the former implicit models, the explicit models can more easily handle structured parts in texture, such as edges and bars, which emerge in high-resolution image textures. However, the computation or detection of such textons is not trivial. It is also worth noticing that modeling the interactions between textons may involve heavy computation.

\subsubsection{Co-occurrence patterns of pixels for texture analysis}
Co-occurrence matrices~\cite{texture_Haralick73,texture_Larry} is still a popular approach for texture analysis. It characterizes image textures with a set of sufficient non-parametric and low-order statistics of pixels.
The preliminary use of co-occurrence matrices involved in statistics of pairwise pixel relationships in several predefined fashions. Local Binary Pattern (LBP)~\cite{pietikainen2000rotation} extended the concept of co-occurrence by developing a framework for studying the statistics of co-occurrent binary patterns.
Though, this kind of approaches demonstrated strong capability to characterize random and near random textures~\cite{pietikainen2000rotation}, one limitation of them is, however, a lack of consideration for large-scale and geometrical structures in texture images.

\subsubsection{Computing textons with high-order statistics}
{
The convolutional network -based methods~\cite{sifre2013rotation, Cimpoi2014, Cimpoi2015} have attracted a lot of attentions recently. In these kind of methods, the cascaded networks are resorted to describe the high-order statistics. More precisely, in~\cite{sifre2013rotation}, a three-layers of rotation, scaling and deformation invariant scattering transform network is designed for texture discrimination. Though it achieves a good performance, it is still difficult to depict the large-scale structure in texture since the network is not deep enough.
Besides, some other novel methods\cite{Cimpoi2014,Cimpoi2015} based on Convolutional Neural Networks(CNN) can describe the local and large-scale structure of textures by using many cascaded covolutional layers and pooling layers in the pre-trained network. However, since the network, e.g. VGG~\cite{Simonyan14c}, is trained on ImageNet rather than a texture dataset, it is hard to explain how the extracted feature maps define the characteristic of textures, especially the invariance attributes of scale, rotation and illumination.}

\subsection{Contributions}
As mentioned before, this paper tries to combine the shape-based texture analysis scheme and the co-occurrence patterns methods in the texton-based paradigm.
It can inherit, from the shape-based texture method, both the strong ability to model the geometrical aspects of textures and the high robustness to imaging conditions changes, and also provides a more flexible way to consider high-order statistics on the tree by investigating the co-occurrence patterns of shapes.

The main contribution of this paper is to present an efficient and flexible shape-based texture analysis framework by investigating the co-occurrence patterns of shapes.
More precisely, as illustrated in Figure~\ref{fig:flowchart}, given a texture, we first decompose it into a tree of shapes relying on a fast level set transformation~\cite{monasse2000fast}, where each shape is associated with some attributes. We then learn a set of co-occurrence patterns of shapes from a set of texture images, \emph{e.g.} by K-means algorithm and others. Taking the learned shape co-occurrence patterns as visual words, a bag-of-words model is finally established to describe a texture image. In contrast with SITA~\cite{xia2010shape}, the proposed method provides a more flexible way to consider more complex shape relationships and high-order statistics on the tree.
Moreover, as we shall see, SITA can be regarded as an special case of the proposed one, that only marginal distributions and simple statistics of pair shapes are taken into account. To our knowledge, this is the first time to use co-occurrence patterns of explicit shape for texture analysis.
Several experiments of texture retrieval and classification demonstrate the efficiency of the proposed analysis method on various datasets.

In the rest of the paper, we briefly introduce the background and recall the shape-based texture analysis in Section~\ref{SEC:pre}. In Section~\ref{SEC:SCOP} we present the proposed SCOP texture analysis method in detail. In Section~\ref{SEC:exp}, the ability of these features to classify or retrieve textures is demonstrated on various datasets. Section~\ref{SEC:con} concludes the paper with some remarks.
A preliminary version of this paper has been presented in~\cite{xia2014texture}.

\section{Preliminary}
\label{SEC:pre}

This section recalls the preliminary of our work, {\em i.e.}, tree-of-shapes (ToS) image representation, moment-based shape description and shape-based invariant texture analysis.
\subsection{Tree-of-shapes (ToS) image representation}
Given a gray-scale image $u: \Omega \mapsto G $ with $ G = \{0,\cdots\,g-1 \}$ and $\Omega=\{0,1,\cdots, n-1\} \times \{0,1,\cdots, m-1\}$, its upper and lower level sets, denoted by $\{\chi_\lambda(u)\}_{\lambda}$ and $\{\chi^\lambda(u)\}_{\lambda}$, are respectively defined as
$$\chi_\lambda(u)= \{ x \in \Omega; \ u(x) \geq \lambda \}$$
and
$$\chi^{\lambda}(u)= \{ x \in \Omega; \ u(x) \leq \lambda \},$$
for $\lambda \in G$.

Note that $\{\chi_\lambda(u)\}_{\lambda}$ (respect. $\{\chi^\lambda(u)\}_{\lambda}$) is a complete image representation, from which the image $u$ can be reconstructed without losing any information,
$$
u(x) = \min\{\lambda; \, x \in \chi_\lambda(u), \forall \lambda \}.
$$
Another interesting property of the level sets is $\chi_{\lambda_1}(u) \subset \chi_{\lambda_2}(u)$ and $\chi^{\lambda_2}(u) \subset \chi^{\lambda_1}(u)$ if the gray level ${\lambda_1} < {\lambda_2}$, which implies that the connected components of upper/lower level stes are naturally embedded in a tree structure~\cite{caselles1999topographic}.
In~\cite{monasse2000fast}, it has reported that these two redundant tree structures can be combined to derive a hierarchical representation of images, named {\em tree-of-shapes} or {\em topographic map}, by drawing a notion of {\it shape}.

\begin{myDef}[\textbf{Shape}]
A {shape} $s$ is a connected component of upper or lower level sets, whose holes have been filled with pixels. It corresponds to the interior of a level line (the boundary of a level set).
\end{myDef}

\begin{figure}[ht!]
\centering
  \includegraphics[width= 0.6\linewidth]{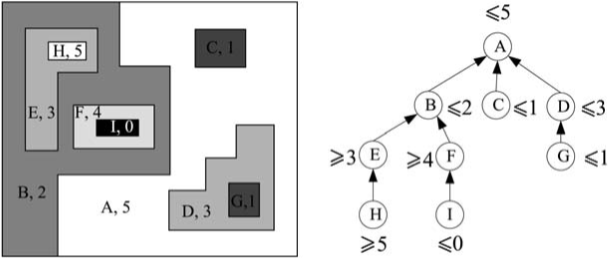}
\caption{Representation of a synthetic image by its topographic map. Left: an original digital image; Right: representation of the image by its tree of shapes, where
$(A,B,\dots,I)$ denote the corresponding shapes.}
\label{fig:flst}
\end{figure}

The tree-of-shapes of an image can be computed by an efficient algorithm, \emph{i.e.} Fast Level Set Transform (FLST)~\cite{monasse2000fast}.
Figure~\ref{fig:flst} shows an example of the tree-of-shapes representation of a synthetic image. Shapes in this example are $\{A, B, \cdots, I\}$, polygons without holes.

Based on the tree-of-shape representation, one can define the smallest shape of pixel as follows:
\begin{myDef}[\textbf{The smallest shape of pixel}]
For a pixel $x \in \Omega$, the shape
$s_x = \arg \min_{s} \{|s|; x \in s, \forall s\}$
is called its smallest shape, which is the shape containing $x$ and with the smallest area.
\end{myDef}

Remark that a shape $s$ is a point set and for any two given shapes $s_i$ and $s_j$ with $|s_i| \leq |s_j|$, there is only two kind of relationships: $s_i \subseteq s_j$ or $s_i \cap s_j = \varnothing $.
More details on the tree-of-shape image representation can be found in~\cite{Monasse08}.

As the tree of shapes provides a complete representation of images, it is natural to argue that the modeling of a texture $u$ is then converted to model the tree of shapes, {\em i.e.},
$$
u = \langle V, E \rangle,
$$
where $V = \{s_i\}$ denotes shapes, vertexes on the ToS, and $E = \{e_{ij}\}$ indicates the edges between the shapes $s_i$ and $s_j$, with $e_{ij} = \langle s_i, s_j\rangle$.

\subsection{Moments-based shape description}
The $(p+q)$-th order central moments of shape $s:=\{x; x=(x_1, x_2) \in \Omega\}$ is defined as
\begin{align}
\mu_{pq}(s)=\int\int_s(x_1-\bar{x}_1)^p(x_2-\bar{x}_2)^q\; dx_1 dx_2,
\end{align}
where $(\bar{x}_1, \bar{x}_2)$ is the center of mass of $s$, \textit{i.e.}
\begin{align}
\qquad \bar{x}_i =\frac{1}{\mu_{00}(s)}\int\int_s x_i\; dx_1 dx_2, \,\, \forall i = 1, 2.
\label{eq:centerxy}
\end{align}
where $\mu_{00}(s)$ is the area of the shape $s$.

Considering the similarity (scaling and rotating) invariance of shapes, it thus turns to use the normalized inertia matrix
to derive the shape attributes.
Denote $\lambda_1$ and $\lambda_2$ as the two eigenvalues of the normalized inertia matrix of $s$, with $\lambda_1 \geq \lambda_2$. $\mu_{00}(s)$ is in fact the area of shape $s$. Let $p$ be the shape's perimeter,
the geometrical attributes used for characterizing shape $s$ are then given in Table~\ref{tab:sa}.
Notice that, in what follows for the simplicity, the index of shape $s$ is omitted for each attribute.

\begin{table}[ht!]
\centering
  \caption{Attributes for characterizing a shape $s$. For the notation of symbols, please refer to the text for details.}
  \begin{tabular}{r|l}
  \hline
  Attribute & Computation \\
  \hline
     Elongation&  $\epsilon = \lambda_2 / \lambda_1$ \\
    Ellipse-compactness& $\kappa_e = 1 /(4\pi\sqrt{\lambda_1\lambda_2})$  \\
    Circle-compactness& $\kappa_c = 4\pi p^2 / \mu_{00} $ \\
    \hline
  \end{tabular}
  \label{tab:sa}
\end{table}

\subsection{Shape-based invariant texture analysis (SITA)}
Based on the above mentioned arguments, Xia {\em et al.}~\cite{xia2010shape} introduced a shape-based texture analysis scheme.
In order to characterize the vertex $V = \{s_i\}$ of the ToS, one can compute moments-based attributes to describe the geometrical aspects of each shape.


In order to introduce some radiometrical information of shapes, \emph{e.g.} contrast, which are reported to be important for texture discrimination, the smallest-shape-normalized contrast is used, \emph{i.e.}
$$ \gamma = \frac{u(s)- \mu_s}{\sigma_s}, $$
where $u(s)$ indicates the gray value of the pixels only contained by $s$, $\mu_s$ and $\sigma_s$ as the mean and the standard deviation of gray values of all pixels inside $s$.

To derive shape attributes that are robust to affine transformations, SITA proposed to calculate the ratio between the area of shape $s$ and the average area of its $r$-order ancestor family $s^r, r\in[1,\cdots,M]$,
$$
\beta= M a / (\sum_{r=1}^M a(s^r)).
$$

Moreover, observe that the shapes extracted from the upper level sets correspond to bright poles in images, while shapes extracted from the lower level sets correspond to dark ones. It thus proposed to add a \emph{Polarization} attribute
$$
\phi(s) =
\begin{cases}
  + & s \in \chi_\lambda\\
  - & s \in \chi^\lambda
\end{cases}
$$
to each shape $s$ for distinguishing this information.

Thus, as a summarization, the attributes of shape $s$ can be written as,
\begin{align}
f(s) = (\epsilon, \kappa_e, \kappa_c, \gamma, \beta)^T \cdot \phi(s).
\label{eq:attri}
\end{align}

The marginal distributions of shape attributes are used for texture characterization, which actually assumes that the shapes and their attributes are independent on the ToS and only derive statistics from the vertexes $V$ for texture analysis. However, high-order statistics on ToS, contained in the edges $E$ plays an important role for texture analysis. The SITA derived this information simply using the scale ratio $\beta$ in Eq.\eqref{eq:attri}, which is not enough to describe the high-order statistics.

As we shall see, the following part of this paper proposes a general and flexible way for considering more comprehensive high-order statistics on ToS.

\section{Texture modeling with SCOPs}
\label{SEC:SCOP}

\subsection{Texture modeling with explicit shape textons}

According to the texton theory~\cite{Julesz}, shapes $s$ can be regarded as textons, the the basic elements for texture analysis. While, compared with existing ones, shapes textons are more flexible to describe the geometrical aspect of textures.
In order to derive a texture model from the ToS, one needs to estimate the arrangement rules of the shape textons.
In this paper, we argue that a texture image can be characterized by the distribution of shapes on the ToS statistically.
Actually, the ToS-based texture model has following nice properties :
\begin{itemize}
\item[1)] It enables texture analysis at several scales simultaneously without geometric degradation when going from fine to coarse scales, as the ToS representation provides a non-linear scale-space.
\item[2)] It is invariant to increasing contrast changes to texture images.
\item[3)] It uses shapes as local textons for texture analysis and thus leads to a flexible way to achieve geometrical invariance by normalizing the geometrical transforms from shapes individually.
\end{itemize}

\begin{figure}[ht!]
\centering
\subfigure[two highly structured textures]{
    \includegraphics[width=0.3\linewidth]{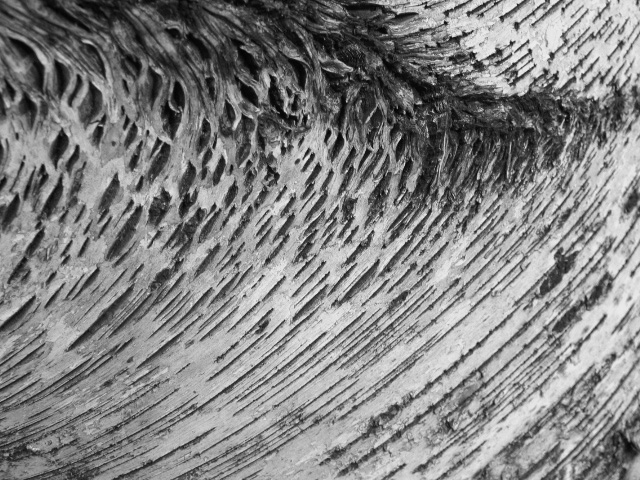}
    \includegraphics[width=0.3\linewidth]{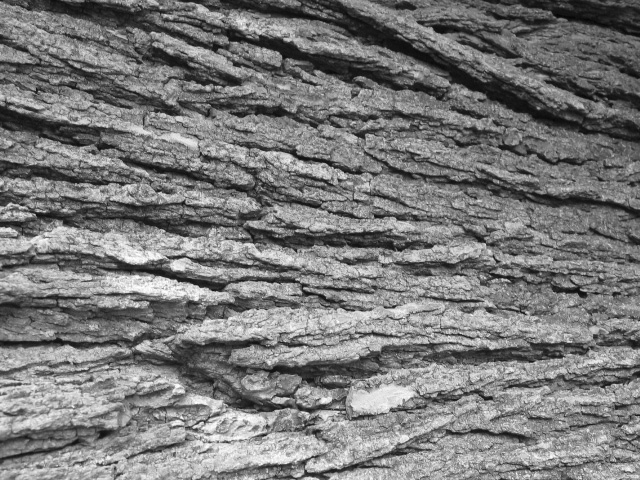}}
\subfigure[the shape-based representations of the textures in (a)]{
    \includegraphics[width=0.3\linewidth]{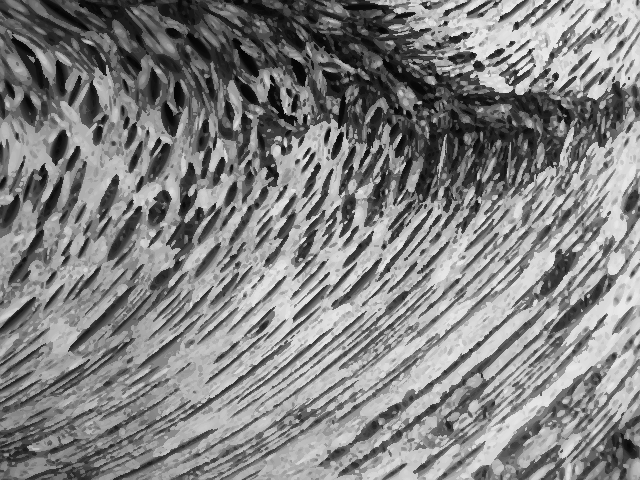}
    \includegraphics[width=0.3\linewidth]{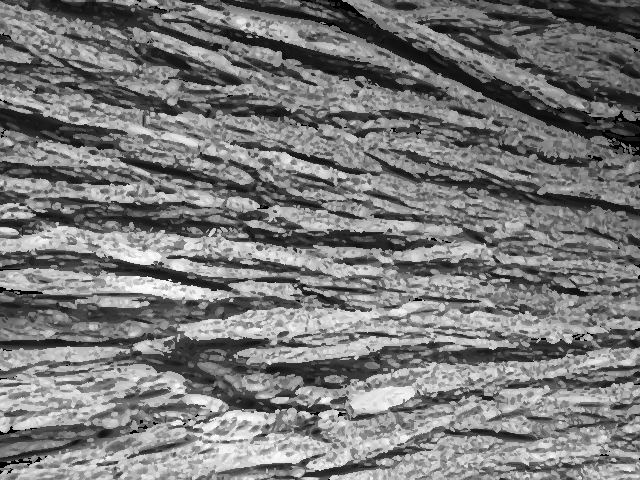}}
\caption{Illustration of textures represented by second-order shapes, \emph{i.e.} ellipses. Two original textures are shown in (a) and the corresponding approximated texture is shown in (b), which keeps the hierarchy of the tree but replaces each shape by an ellipse with the same second-order statistics. We can see that the elongated structures are well represented.}
\label{fig:shaperec}
\end{figure}

\textbf{{\em Why does SITA work}?} It is now worth rechecking the reason why SITA~\cite{xia2010shape} works so well.
SITA actually follows the texton-based texture analysis scheme, by regarding the explicit shapes as textons in the paradigm.
As the analysis scheme use explicit shapes to characterize textures, it demonstrated strong ability to depict the geometrical aspects of texture images, and thanks to the flexible normalization of geometrical transformations from individual shapes, it showed high efficiency for achieving geometric invariant texture features and reported superior performance on invariant texture recognition task.
In particular, it proposed to use second-order moments to describe shapes on the tree, which actually implies that each shape is approximated by an ellipse. This efficiency can be observed from Fig.~\ref{fig:shaperec}, where textures are reconstructed by keeping the tree structures while replacing each shape on the tree by an ellipse with the same second-order moments.
Observe that the original image and the image approximated by ellipses are hard to distinguish.

However, the SITA scheme simply assumed that the shape attributes are independent and modeled textures by several marginal distributions. One limitation of such a scheme lies in its difficulty to handle the relationships of shapes, corresponding to high-order statistics on the tree, which is reported to be important for texture discriminations, see~\cite{xia2010shape} for more details, where only the scale ratio between pair of shapes were considered and performed better than other attributes.

\subsection{Shape Co-Occurrence Patterns (SCOPs)}
In spatial domain, the relationship of shapes corresponds to local co-occurrence structures in images. In the topographic map, a hierarchical tree, the main relationships are sibling and conclusion, corresponding to small branches on the tree. If we take such small branches as textons, texture modeling is then to investigate the branches arrangement laws on the tree.

In our context, we define \emph{shape co-occurrence patterns} as \emph{simple and local common branches on the tree of shapes}, which reflects specific spatial organization in the topographic map.
Observe that, though the shape co-occurrence patterns are local on the tree of shapes, they may correspond to large spatial areas of pixels in the images, which enables us to consider highly geometrical and complex texture elements, such as co-occurred elongated or sharp edges.

Denoting $s^r$ as the $r$-order ancestor of shape $s$ on the ToS, $s'$ as the sibling of $s$ and $s^{\tau}$ to be the ${\tau}$-order grand-ancestor of $s$ with $\tau > r$, in this work, we define following four kind of SCOPs as
\begin{align}
    \textrm{single shape} \,(\textrm{SS}): & \, s, \,  \nonumber \\
    \textrm{shape-ancestor} \,(\textrm{SA}): & \, s \rightarrow s^r, \,    \nonumber\\
    \textrm{shape-ancestor-grandancestor} \,(\textrm{SAG}): & \, s \rightarrow s^r \DashedArrow[->,densely dashed    ]  s^{\tau}, \,   \nonumber\\
    \textrm{shape-ancestor-sibling} \,(\textrm{SAS}): & \, s \rightarrow s^r \leftarrow s^{\prime}, \, \nonumber
\end{align}
respectively. Note that the shape-ancestor-grandancestor (\textrm{SAG}) can be extended into a higher layer structure by a lager parameter $\tau$.

\begin{figure}[htb!]
  \centering
    \includegraphics[width= 0.5\linewidth]{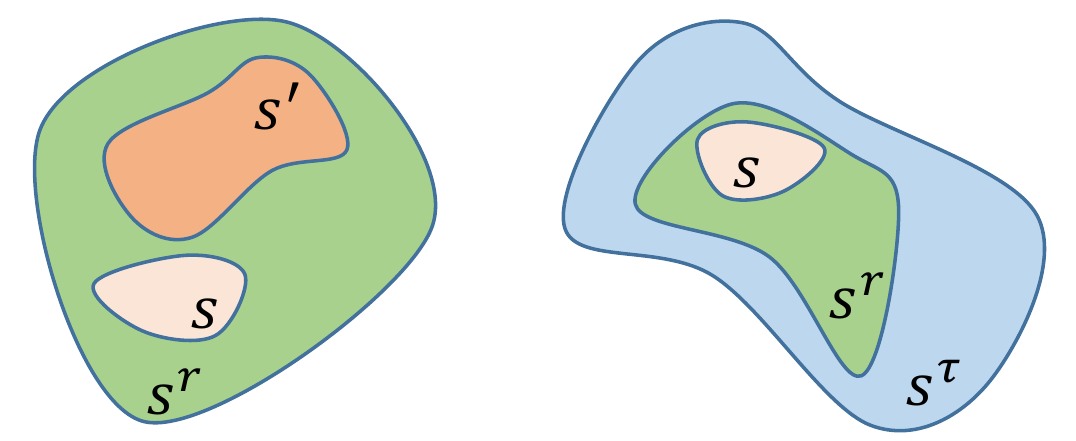}
\caption{The shape-ancestor-sibling $(SAS: \, s \rightarrow s^r \leftarrow s^{\prime})$ (a) and shape-ancestor-grandancestor $SAG: \, s \rightarrow s^r \rightarrow s^{\tau}$ (b).}
  \label{fig:pattern}
\end{figure}

Fig.\ref{fig:pattern} illustrates two common SCOPs $SAG$ and $SAS$, which are very popular and efficient in texture presentation as they encode rich spatial co-occurrence information.

Two parameters for constructing the SCOPs are the interval order $r$ between the shape and its ancestor, and the levels of the interval $\frac{\tau}{r}$.

\subsubsection{Estimating the interval $r$}
The interval $r$ controls the correlation between two adjacent shapes in the tree.  Observe that a too small $r$ makes no sense as the two adjacent shapes will be almost the same, while a too big $r$ will not be local enough and results in weak descriptive ability.
In our context, we proposed to use $r$, that makes the average difference between the areas of the pair of shapes larger than the perimeter $p$ of shape $s$,

\begin{align}
r = \min  \{i; \, a_{s^i} - a_{s} > p(s)\},
\label{eq:areaper}
\end{align}
where $s^i$ denotes the $i$-th order ancestor of the shape $s$. $a, p$ denote the area and perimeter respectively.

Note that $r$ in fact approximates the number of blur around objects in nature images. The final choice of the interval $r$ is the average value given by Eq.~\eqref{eq:areaper} on all shapes in the images of the dataset, which corresponds to the averaging blur in textures.

\subsubsection{Estimating the number of cascaded ancestors $\tau$}
$\tau$ has similar effect for texture description as $r$. Considering the casual relationship between two cascaded shapes in the tree structure, $\tau$ is set to be integer times of $r$. In our experiments, $\tau$ is set to be $2r$ to achieve better performances.

\subsection{Encoding SCOPs for texture analysis}
As a local common branch on the tree, each SCOP actually contains tremendous image or shape realizations described by branch attributes. For the common SCOPs, the attributes are
\begin{align}
    \textrm{descriptor of SCOP-SS}: & \, f(s), \,  \nonumber  \\
    \textrm{descriptor of SCOP-SA}: & \, [f(s), f(s^r)] \,  \nonumber  \\
    \textrm{descriptor of SCOP-SAG}: & \, [f(s), f(s^r), f(s^{\tau})], \,  \nonumber \\
    \textrm{descriptor of SCOP-SAS}: & \, [f(s), f(s^r), f(s^{\prime})], \, \nonumber
\end{align}
where $f(s)$ is the attribute associated to $s$， which is written in Eq.~\eqref{eq:attri}.

In order to simplify the problem, we need to quantize the shape realizations of each SCOP into limited number of clusters, named \emph{words} of a SCOP. Such \emph{words} can be learned from the tree of shapes of a given set of texture images, as illustrated in Figure~\ref{fig:flowchart}. In this Figure, the different patterns of shapes are extracted from the given dataset and the corresponding \emph{words} are learned respectively. With each \emph{word} of the pattern, any texture image can be encoded into a specific descriptor.

\begin{figure}[!htb]
	\centering
	\includegraphics[width=0.7\linewidth]{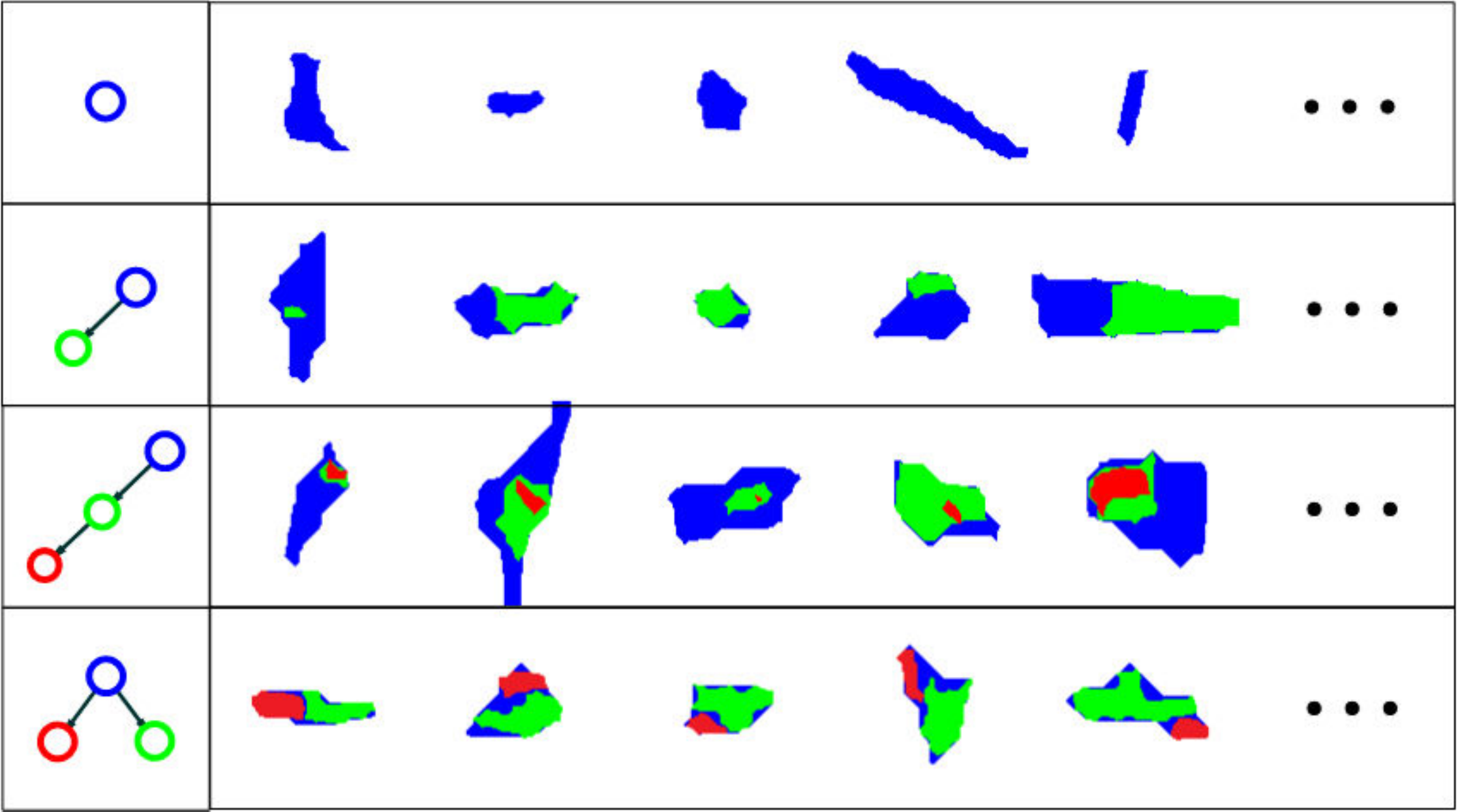}
	\caption{4 SCOPs with several learned \emph{words} by K-Means in corresponding row, $\tau=2r$.}
	\label{fig:words}
\end{figure}

We investigate three kinds of coding methods in the following, which are K-Means, Sparse Coding and Fisher Coding method. Figure~\ref{fig:words} shows several typical learned \emph{words}k with K-Means for each SCOP.

\subsubsection{KM-SCOP}
K-Means Coding is one of hard-voting strategy. For each pattern in SCOP, codewords $D\in \Re^{h\times K}$ are build by clustering the shape feature with K-Means method, where $h$ is the dimension of the SCOP features and $K$ is the number of atoms in the dictionary. Fig.~\ref{fig:words} illustrates the typical kinds of codewords of different patterns, in which shape relationships are hierarchical and coordinate.

After the codewords are obtained, the probability density distributions of shape are represented by histogram, whose bins reflect the occurrence frequency of shape features. The $k$-th bin of the histogram of image $I$ on each pattern can be calculated by
\begin{align}
H(k) = \sum_i \delta(k, \arg\min_{\hat{k}} \| \bm{f} (s_i)-D (\hat{k})\|),\quad k=1,\cdots,K,
\label{eq:hist_km}
\end{align}
where $i$ denotes the index of shape in the topographical map of image $I$, $\delta(x_1,x_2)$ equals $1$ if $x_1=x_2$ and $0$ otherwise.

\begin{figure*}[ht!]
\centering
\includegraphics[width=0.9\linewidth]{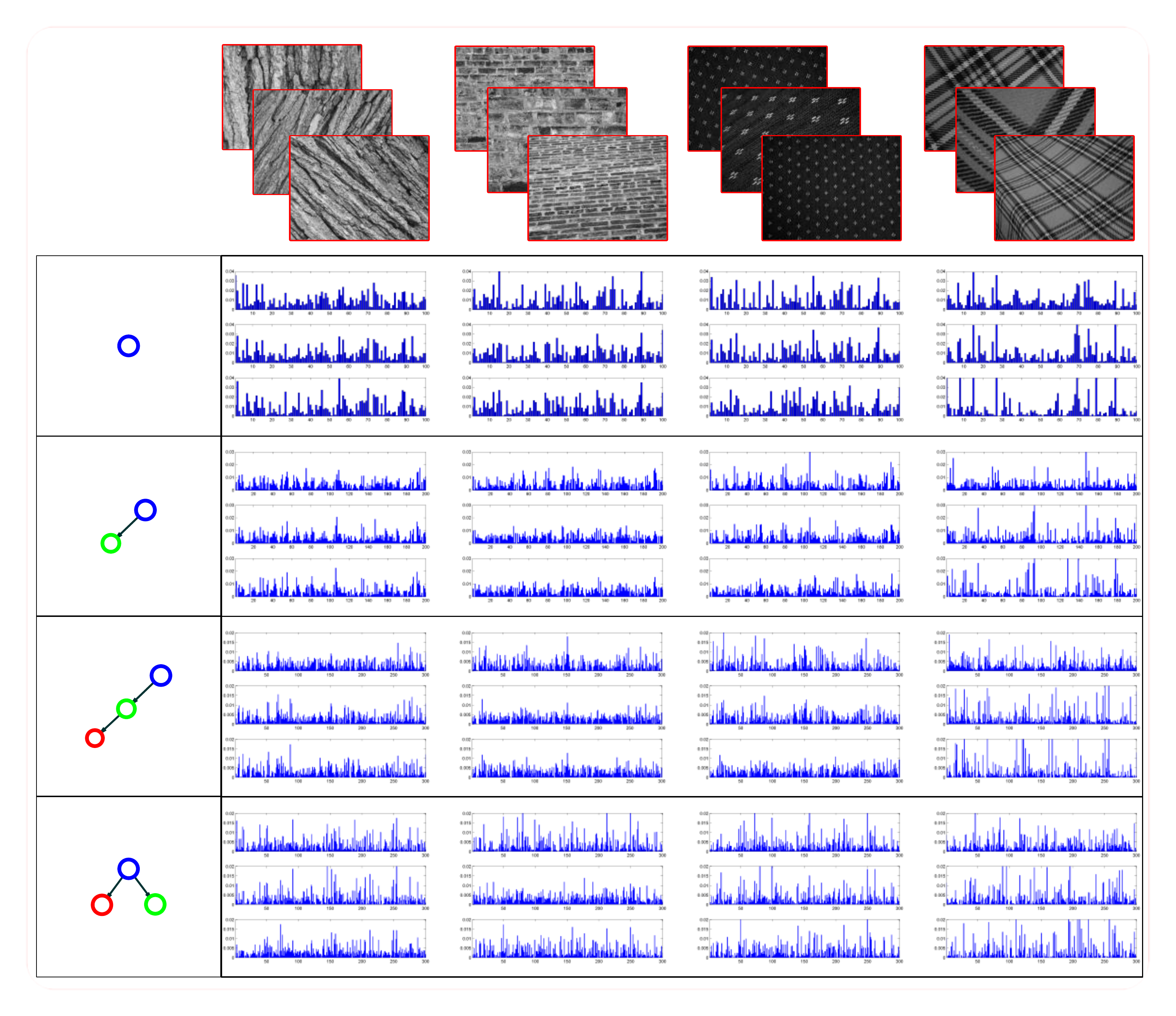}
\caption{The histogram corresponding to different patterns by $12$ texture images containing $4$ different classes with $3$ images per class from UIUC dataset.}
\label{fig:histall}
\end{figure*}

Each histogram is obtained with the corresponding pattern and codeword. Fig.~\ref{fig:histall} shows the example of histograms of typical patterns by $12$ texture images chosen from $4$ different classes with $3$ images per class in UIUC dataset. It is apparent that the histograms of the images belong to the same class are similar while the histograms of the images belong to different classes are different, which means all the histograms are distinguished features and can be used to describe the texture images. Finally, we cascade these histograms as to be the descriptive feature of each texture image by
\begin{align}
\varpi = [H_+^1, H_-^1, H_+^2, \cdots, H_+^Q, H_-^Q].
\label{eq:hist_kmall}
\end{align}
where $Q$ is the total number of patterns.

\subsubsection{SC-SCOP}
This sparse coding method is inspired by \cite{ren2013histograms}, in which histogram of sparse coding is used to be the descriptor of images. In fact, this method could be regarded as a soft-voting strategy. In this strategy, shape features are represented by a combination of several atoms in codewords with different weights, while the weight can be learned by a optimization function.

For building the sparse codewords of each pattern, we select the training feature set $x=\{\bm{f}(s_i\}_{i=\{1,2,\cdots,n_d\}}$ randomly from the dataset, where $n_d$ is the number of samples used for training a codewords. The classic $\ell_1$ sparse coding method~\cite{mairal2009online}, also called \emph{Lasso Optimization}, could be written as
\begin{align}
\ell(x,D)\triangleq \min_{\alpha \in \Re^K} \frac{1}{2} \|x-D\alpha\|^2_2 + \lambda\|\alpha\|_1
\label{eq:sc}
\end{align}
where $\lambda$ is called regularization parameter, $D\in \Re^{h\times K}$ denotes the sparse dictionary and $\alpha$ denotes the corresponding sparse coding matrix of the SCOP features set.

After the codewords are obtained, the sparse coding vector $\alpha_i$ of feature of shape $s_i$ could be calculated by solving
\begin{align}
\min_{\bm{\alpha}_i \in \Re^K}
\| \bm{f}(s_i)
-\bm{D}\bm{\alpha}_i \|^2_2,  \emph{s.t.}
\|
\bm{\alpha}_i
\|_1\leq\lambda,
\label{eq:lasso}
\end{align}
where $i$ denotes the index of shapes, $f(s_i)$ denotes the shape feature maps. Eq.\eqref{eq:sc} and Eq.\eqref{eq:lasso} are solved by Sparse Modeling Software (SPAMS)~\cite{mairal2014sparse}.

As sparse coding represents the data efficiently as a combination of a few typical atoms from the codewords by the weight vector $\bm{\alpha}$, the statistical histogram of sparse coding for each pattern of the images is obtained by~\cite{ren2013histograms}
\begin{align}
H = \sum_i |\bm{\alpha}_i|
\label{eq:hist}
\end{align}
where $i$ denotes the index of shape in the topographic map of image $I$. Finally the SCOP descriptor is the combination of all histograms as Eq.\eqref{eq:hist_kmall}.

\subsubsection{FC-SCOP}
In fisher coding method, the probability density distribution of shape feature is described by the Gaussian Mixture Model (GMM), whose parameters $\Phi=(\mu_k,\Sigma_k, \pi_k: k=1,\cdots, K)$, denoting the mean vector, covariance matrix and the weight of the Gaussian distribution, can be estimated by the expectation maximization (EM) algorithm~\cite{bailey1994fitting}.

For each shape feature $\{f_i\}_{i\in \{1,\cdots,N\}}$ extracted from image $I$, the GMM associates them to a mode $k$ in the mixture with a strength given by the posterior probability
\begin{align}
q_{ik} = \frac{\exp\left(-\frac{1}{2}(f_i-\mu_k)^T \Sigma_k^{-1}(f_i-\mu_k)\right)}
              {\sum_{j=1}^K \exp\left(-\frac{1}{2}(f_i-\mu_k)^T \Sigma_k^{-1}(f_i-\mu_k)\right)}.
\label{eq:gmmprob}
\end{align}

Then we can estimate the feature parameters about mean and covariance deviation vectors for the texture $I$ for each mode $k$ on GMM with
\begin{align}
u_{lk} &=\frac{1}{N\sqrt{\pi_k}}\sum_{i=1}^N q_{ik}\frac{f_{li}-\mu_{lk}}{\sigma_{lk}},\\
v_{lk} &=\frac{1}{N\sqrt{2\pi_k}}\sum_{i=1}^N q_{ik}\left(\left(\frac{f_{li}-\mu_{lk}}{\sigma_{lk}}\right)^2-1\right),
\label{eq:gmmpara}
\end{align}
where $l$ denotes the index of shape attribute feature in Eq.\eqref{eq:attri}. According to this, the fisher coding of each pattern can be written as
\begin{align}
  H=(u_1,u_2,\cdots, u_K,v_1,\cdots, v_K)^T.
\label{eq:gmmh}
\end{align}
The dimension of FC-SCOP is too complex to process in our classification, so we reduce its dimension by PCA\cite{JEGOU-2011-633013}. Following this procedure, Eq.\eqref{eq:hist_kmall} give the final descriptor of a texture image.

Algorithm~\ref{alg:SCOP} provides the implementation pipeline of our methods.
\begin{algorithm}[H] 
 \caption{Image Classification with SCOP}
 \begin{algorithmic}[1]
 \Require  {Train image sets $I_{Tr}$, Test image set $I_{Te}$, dictionaries sizes $\{K^t, t=1, 2, \cdots,Q\}$, minimal shape area $a_{min}$ and maximal shape area $a_{max}$.}
 \Ensure  {Classification result of test image set}
  \State  Calculate the interval $r$ with Eq.~\eqref{eq:areaper}.
  \State  Calculate all the SCOP $\{\textbf{f}^t, t=1, 2, \cdots,Q\}$ of train image sets with Eq.~\eqref{eq:attri};
  \State  Build codewords with $\{\textbf{f}_\pm^t, t=1, 2, \cdots,Q\}$ and the corresponding coding features via K-means coding, sparse coding or Fisher coding method.
  \State  Calculating all the histograms $H$ of the images $I$ in train sets and test sets with  Eq.~\eqref{eq:hist_kmall}.
  \State  Train a SVM classifier $\digamma$ by libSVM~\cite{chang2011libsvm} with the HIK or RBF kernel.
  \State  Classify the test set images with the SVM classifier $\digamma$.
 \end{algorithmic}
  \label{alg:SCOP}
\end{algorithm}

\section{Experimental Evaluations}
\label{SEC:exp}

\subsection{Experiment setup}
We test our approach on two different tasks on which texture information dominates: invariant texture recognition and scene classification.
For the invariant texture recognition task, we test the proposed method on three common used texture datasets, UIUC~\cite{lazebnik2005sparse}, UMD~\cite{xu2009viewpoint}, Brodaz~\cite{brodatz1966textures}. We compare our methods with Scattering transform(ST)~\cite{sifre2013rotation}(With operators of 'log, scale average, multi-scale train') and some conventional  methods~\cite{crosier2008texture,xu2010new,xia2011texture,liu2012} as well as the  deep-learning based method~\cite{Cimpoi2015}. For the scene classification, the proposed method is tested on the MIT Outdoor Scene dataset~\cite{oliva2001modeling}, UC Merced Land Use dataset~\cite{yang2010bag} and WHU-RS19 dataset~\cite{xia2010structural} to evaluate its efficiency for different kinds of datasets. We compare our results with classical methods of Lazebnik~\cite{Lazebnik2006} and Xia~\cite{xia2011texture}, we also test ST~\cite{sifre2013rotation} on these datasets. All the compared methods are tuned to use the suitable parameters.

Our parameters are set as: the minimal shape area is set to be $3$ pixels, the maximal shape area is set to be $0.05$ times of the image size, the size of the codewords of the four patterns are $K=\{100,200,300,300\}$ separately, the regularization parameter $\lambda$ for sparse coding is $0.05$. The reduced descriptor dimension for each pattern in FC-SCOP is $500$. Specificity, Histogram Intersection Kernel (HIK) is used to train the SVM for KM-SCOP, while Radial Basis Function kernel (RBF) is used for SC-SCOP and FC-SCOP. The HIK and RBF kernel are defined as
\begin{align}
I(\varpi^X, \varpi^Y)&=\sum_{k=1} \min(\varpi^X(k),\varpi^Y(k)),\\
R(\varpi^X, \varpi^Y)&=\exp\left(-\frac{\|(\varpi^X)^\mu-(\varpi^Y)^\mu\|_2^2}{2\sigma^2}\right),
\end{align}
where $i$ is the index of the bins of the histogram, $\sigma$ is the scale parameters of the kernel and $\mu$ is the power transform parameters, which is usually set to be $0.3$ for sparse coding\cite{ren2013histograms} and $1$ for Fisher coding.

In our experiments, each pattern (cascaded by corresponding positive and negative patterns) is evaluated separately, then the combined SCOPs are tested and compared.

We evaluate the algorithm by classification and retrieval rate. According to~\cite{texture_Lazebnik2}, the retrieval experiments measure how well that an individual sample is used to model the texture class. But the classification experiments can consider the variance in class and model the texture class with several more samples from the class. The classification accuracy is computed as:
\begin{align}
\gamma=\mathcal{N}_c/\mathcal{N}_s
\label{eq:acc}
\end{align}
where $\mathcal{N}_c$ denotes the number of samples that are classified correctly and $\mathcal{N}_s$ denotes the number of all samples. In our experiments, over $200$ random splits between training sets and testing sets are taken to calculate the average classification result.

The retrieval experiment consists of using any sample as a query to retrieve the $N_r$ most similar ones in the data set. For evaluation, the average number of correctly retrieved samples (generally called recall) when the query spans the whole dataset is drawn as a function of $N_r$. Inspired by~\cite{xia2010shape}, which reported that the use of geodesic distance can improve the retrieval results, we calculate the geodesic distance for all compared methods.

Besides, according to~\cite{sifre2013rotation}, multi-scale (MS) train helps to improve classification results. More precisely, decompose the original texture images into spatial scale pyramids and extract features from the resized image at each layer in the pyramids. Then collect all the features as a new set of features of the corresponding texture image. At last, the new set of features can be used to recognize the texture image.

\subsection{Experiments on invariant texture recognition}

This section evaluates our method on three common used texture datasets, and compare our results with the popular texture analyzing methods, as well as the state-of-the-art. In our algorithm, parameters are set up according to information of the datasets, which are described below:
\begin{itemize}
  \item[-] \textbf{UIUC dataset}: This dataset~\cite{lazebnik2005sparse} contains 25 texture classes with 40 images of size $640\times480$ per class. The samples are under different scales, viewpoints and rotations. Due to this, the features stable to these changes are able to achieve a better performance. Moreover, the number of cascaded ancestors $\tau/r$ is set to be $2$ and the interval $r$ between parent and children is $5$ by Eq.~\eqref{eq:areaper}. We evaluate the method with $5$, $10$, $20$ samples per class for training, and the rest for testing by the three encodig strategies.
  \item[-] \textbf{UMD dataset}: This dataset~\cite{xu2009viewpoint} contains 25 texture classes with 40 images of size $1280\times960$ per class. $r$ is equal to $5$ by Eq.~\eqref{eq:areaper}. Considering the structures of the samples in several classes in this dataset are arranged randomly, such as the class of trees or vegetables, which means that the SCOP-SAS is not suitable to improve the overall results, we just construct the SCOP by combining SCOP-SS-SA-SAG. Three coding strategies are implemented and we train the kernel SVM with $5$, $10$, $20$ samples per class respectively, while classify the rest and evaluate by Eq.\eqref{eq:acc}.
  \item[-] \textbf{Brodatz dataset}: This dataset~\cite{brodatz1966textures} contains $111$ different kinds of classes with $9$ images of size $213\times 213$ per class. The samples in each class are cut from one big image separately. As a result, the samples in each class are not variant in viewpoint, rotation and scale. Due to the simple structures of the images in this dataset, the complex patterns are not suitable to model it. In this experiment, SCOP-SS and SCOP-SA are combined into SCOP and the interval $r$ is $3$ by Eq.~\eqref{eq:areaper}. What's more, classifying the dataset with $2$, $3$, $4$ samples per class to train separately is applied with the three coding methods.
\end{itemize}

\begin{figure*}[htb!]
\centering
  \includegraphics[width=0.7\linewidth]{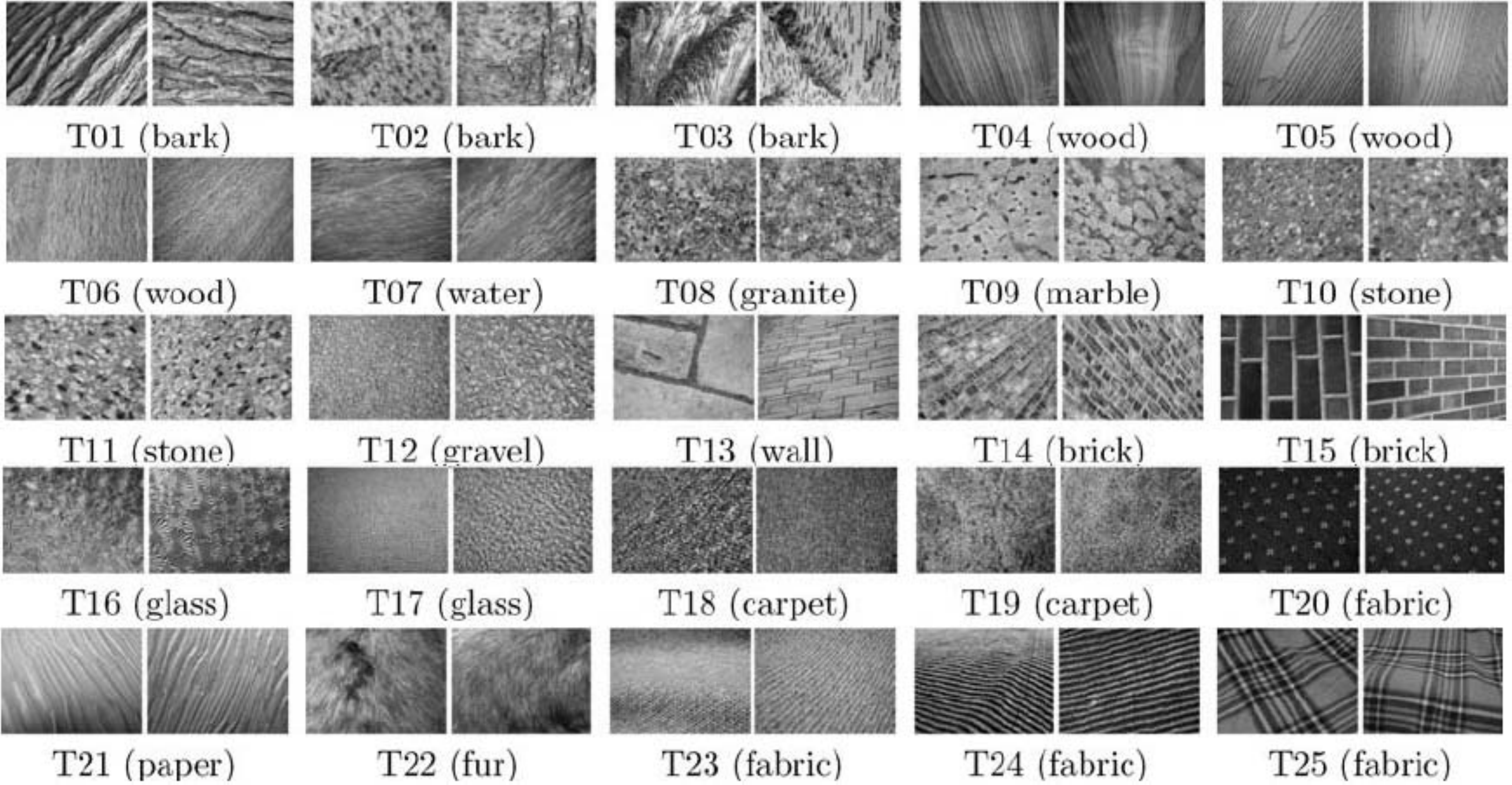}
  \includegraphics[width=0.7\linewidth]{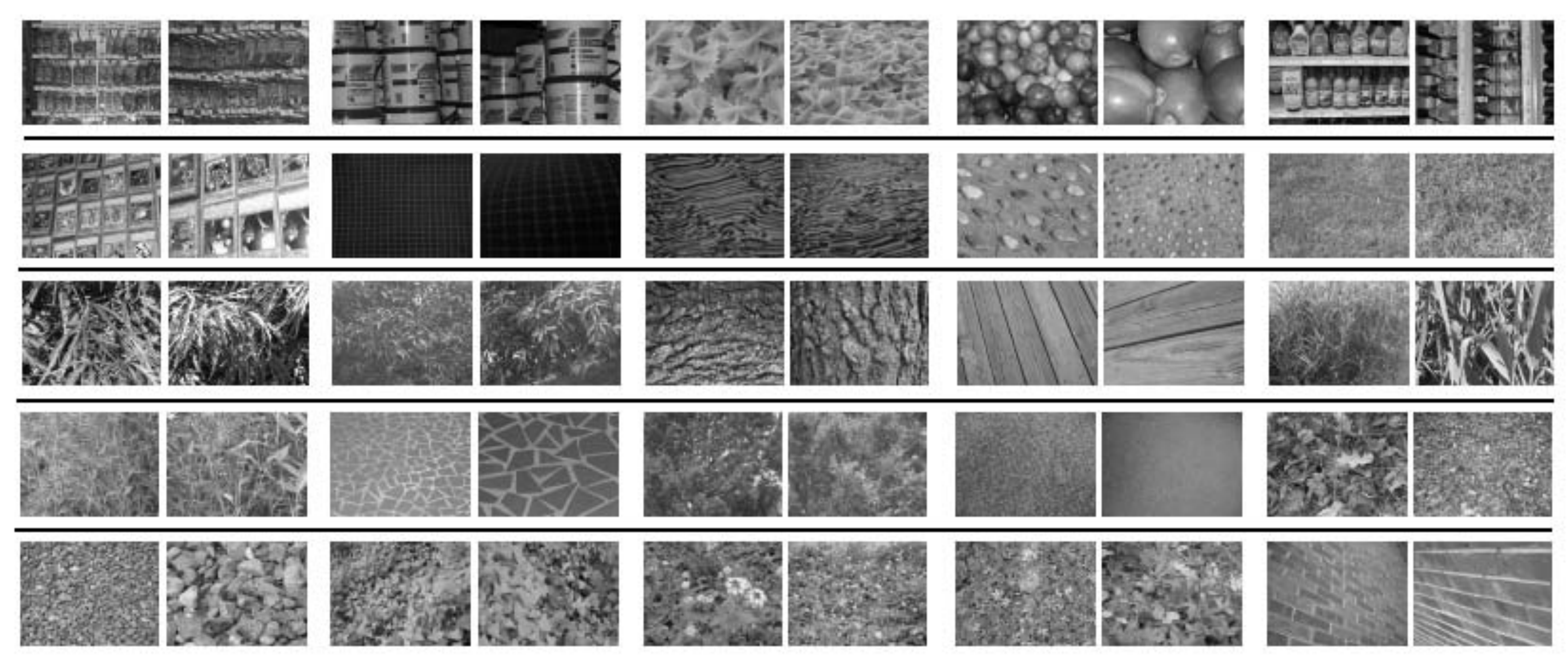}
  \caption{Some samples from UIUC dataset and the UMD dataset}
  \label{fig:invariant_textures}
\end{figure*}

The overall classification results are given in Tab.\ref{tab:texture}. We show the compared results at the top rows, then give the detailed classification results of each pattern with KM-SCOP, finally we present our cascaded patterns by KM-SCOP, SC-SCOP and FC-SCOP methods at the bottom rows. In this table, the state of art, Scattering Transform (ST)\cite{sifre2013rotation} performs very well with some exterior operation, such as logarithmic non-linearity transform and scale average. More precisely, ST outperforms other methods (Including our method without multi-scale train strategy) when the training size is $10$ and $20$ on UMD dataset. Despite this, our algorithm (without multi-scale train strategy) outperforms others in all other cases, especially with the FC-SCOP. Comparing these three coding methods, it is obvious that the FC-SCOP performs best and KM-SCOP method performs worst in these three methods in the case of the classification accuracy. Generally speaking, histogram of K-Means estimates the distribution of shape attribute features with hard-voting strategy, which introduces errors since each shape attribute feature in image is represented by an atom in codewords. The histogram of sparse coding adopts the soft-voting strategy, in which the shape attribute feature is represented by a sparse weight vector. So the accuracy improves from KM-SCOP to SC-SCOP. However, FC-SCOP uses GMM to model the feature sets. In this situation, each shape attribute will be coded by the probability density distribution of GMM with the corresponding parameters. With  FC-SCOP, most of our results outperform the state-of-the-art.

Since the multi-scale training strategy is adopted in~\cite{sifre2013rotation, Cimpoi2015}, here we also test the strategy in our best coding method, FC-SCOP+MS, which is shown in the last row. We can observe a significant improvement according to the table. Although our method is robust to scale changes, it is still intuitive to see that more structures will show up when an image is resized, especially up-sampled. The new structures provide more information to help recognize the textures. The disadvantage is that it cost extra time to extract the whole features.

Besides,  our SCOP-SS  performs better than SITA~\cite{xia2010shape} because we use more attributes of the shapes. Moreover, it is interesting to find that for the texture images with complex structures, the higher-order patterns perform better. For example, in UIUC dataset and UMD dataset, the performance of SCOP-SAG is better than pattern SCOP-SA, and the performance of SCOP-SA is better than pattern SCOP-SS. It proves our intuition that the model of image structures is effectively to depict the images and the more complex structures should be modeled by more complex patterns.

Finally, we can notice that the deep learning method, IFV+DeCAF~\cite{Cimpoi2015}, does not perform well in UIUC dataset and UMD dataset. This is mainly due to the fact that this method resorts to the pre-trained neural network to extract texture features, which does not take into account the complex structure variations, e.g. rotation, scaling and affine deformations, in UIUC dataset and UMD dataset.

\begin{table*}[htb!]
\centering
\caption{Classification results with standard deviations on UIUC~\cite{lazebnik2005sparse}, UMD~\cite{xu2009viewpoint} and Brodaz~\cite{brodatz1966textures} datasets}
\label{tab:texture}
\tiny
\begin{tabular}{c|c|c|c|c|c|c|c|c|c}
  \hline
  & \multicolumn{3}{|c|}{UIUC dataset} & \multicolumn{3}{|c|}{UMD dataset} &\multicolumn{3}{|c}{Brodatz dataset}\\\hline
  Training size                      & 5 & 10 & 20 & 5 & 10 & 20 & 2 & 3 & 4\\\hline
  Lazebnik~\cite{lazebnik2005sparse} & - & 92.6 & 96.0 & - & - & - & - & - & -\\
  WMFS~\cite{xu2010new}              & 93.4 & 97.0 & 98.6 & 93.4 & 97 & 98.7& - & - & -\\
  SITA~\cite{xia2010shape}            & 91.5 & 95.0 & 97.5 & 95.1 & 98.7 & 99.1 & 91.3 & 93.1 & 94.1\\
  ST~\cite{sifre2013rotation}        &93.3$\pm$ 0.4 & 97.8$\pm$ 0.6 & 99.4$\pm$ 0.4 &96.3$\pm$1.0 & 98.9$\pm$ 0.6 & {99.7$\pm$ 0.3} & 86.7$\pm1.3$ & 91.3$\pm$1.1 & 93.9$\pm$1.0\\
  IFV+DeCAF~\cite{Cimpoi2015}           & - & - & 98.96 $\pm$ 0.51 & - & - & 99.52 $\pm$ 0.31 & - & - & - \\
  \hline
  SCOP-SS & 89.8$\pm$ 1.5 & 94.6$\pm$ 0.9 & 97.0$\pm$ 0.7
            & 95.9$\pm$ 1.1 & 98.0$\pm$ 0.7 & 99.1$\pm$ 0.4
            & 92.5$\pm$ 0.8 & 94.6$\pm$ 0.7 & 95.7$\pm$ 0.8\\
  SCOP-SA & 94.2$\pm$ 1.2 & 97.3$\pm$ 0.6 & 98.5$\pm$ 0.5
            & 96.5$\pm$ 1.1 & 98.6$\pm$ 0.6 & 99.4$\pm$ 0.4
            & 91.1$\pm$ 0.9 & 93.6$\pm$ 0.9 & 95.0$\pm$ 0.8\\
  SCOP-SAG & 94.8$\pm$ 1.3 & 97.7$\pm$ 0.6 & 98.7$\pm$ 0.5
            & 96.8$\pm$ 1.0 & 98.7$\pm$ 0.6 & 99.4$\pm$ 0.4
            & 89.0$\pm$ 1.0 & 92.4$\pm$ 0.8 & 94.2$\pm$ 0.9\\
  SCOP-SAS & 93.2$\pm$ 1.2 & 96.1$\pm$ 0.7 & 97.5$\pm$ 0.6
            & 95.1$\pm$ 1.1 & 98.0$\pm$ 0.6 & 99.0$\pm$ 0.4
            & 76.4$\pm$ 1.4 & 81.5$\pm$ 1.1 & 84.5$\pm$ 1.1 \\
  KM-SCOP   & 95.8$\pm$ 1.0 & 97.9$\pm$ 0.5 & 98.9$\pm$ 0.4
            & 97.1$\pm$ 1.0 & 98.7$\pm$ 0.5 & 99.5$\pm$ 0.4
            & 92.6$\pm$ 0.9 & 94.8$\pm$ 0.9 & 95.8$\pm$ 0.7 \\ \hdashline
  SC-SCOP  & \textbf{96.0$\pm$ 1.0} & 98.2$\pm$ 0.6 & 99.3$\pm$ 0.4
                 & 97.1$\pm$1.1 & 98.8$\pm$0.7 & 99.6$\pm$0.4
                 & 94.1$\pm$0.8 & 96.1$\pm$0.7 & 96.9$\pm$0.7\\ \hdashline
  FC-SCOP & 95.7$\pm$1.2 & \textbf{98.5$\pm$0.6} & \textbf{99.5$\pm$0.3}
                & \textbf{97.2$\pm$1.0} & 98.8$\pm$0.6 & 99.6$\pm$0.4
                & \textbf{95.5$\pm$0.6}  & \textbf{97.1$\pm$0.5} & \textbf{97.8$\pm$0.5} \\ \hdashline
  FC-SCOP+ MS & \textbf{97.4$\pm$1.0} & \textbf{99.2$\pm$0.5} & \textbf{99.8$\pm$0.2}
 & \textbf{98.3$\pm$0.8} & \textbf{99.3$\pm$0.5} & \textbf{99.7$\pm$0.2}
 & \textbf{96.7$\pm$0.7}  & \textbf{98.1$\pm$0.6} & \textbf{98.8$\pm$0.4} \\
  \hline
\end{tabular}
\end{table*}

\begin{figure*}[htb!]
\centering
  \includegraphics[width=0.3\linewidth]{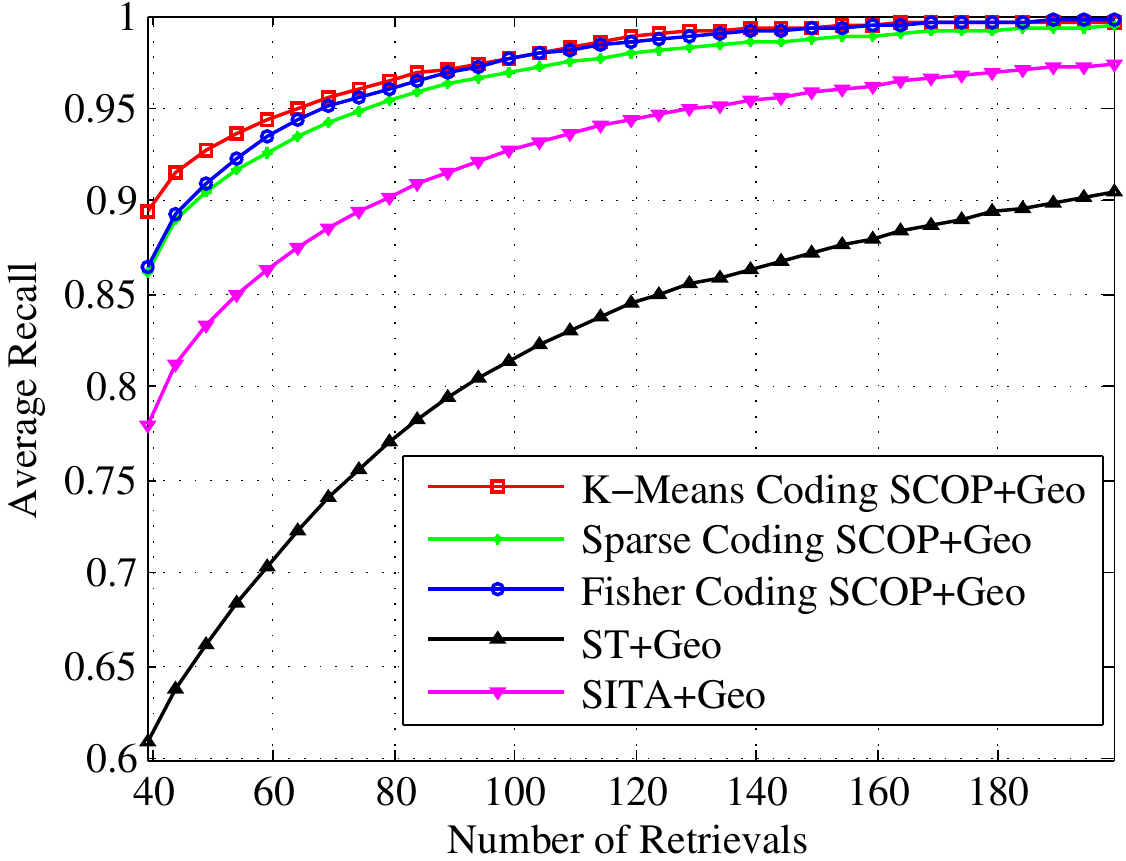}
  \includegraphics[width=0.3\linewidth]{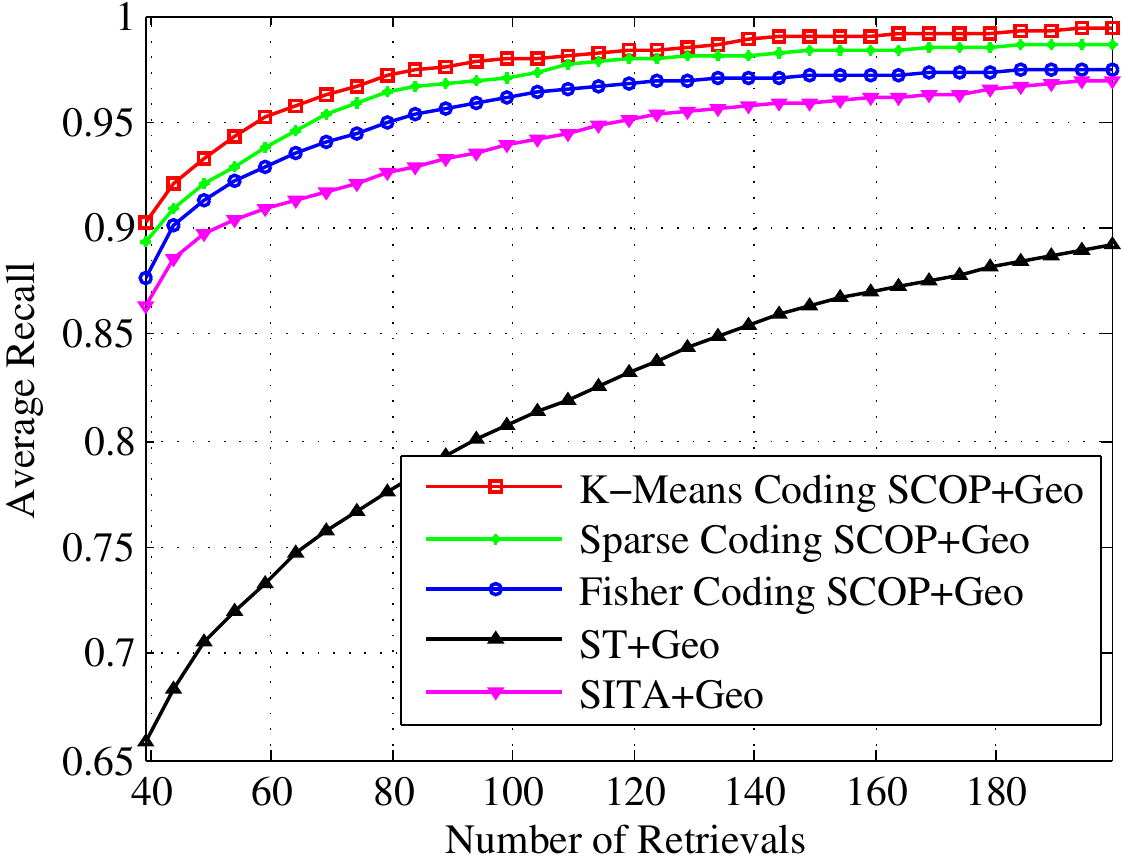}
  \includegraphics[width=0.3\linewidth]{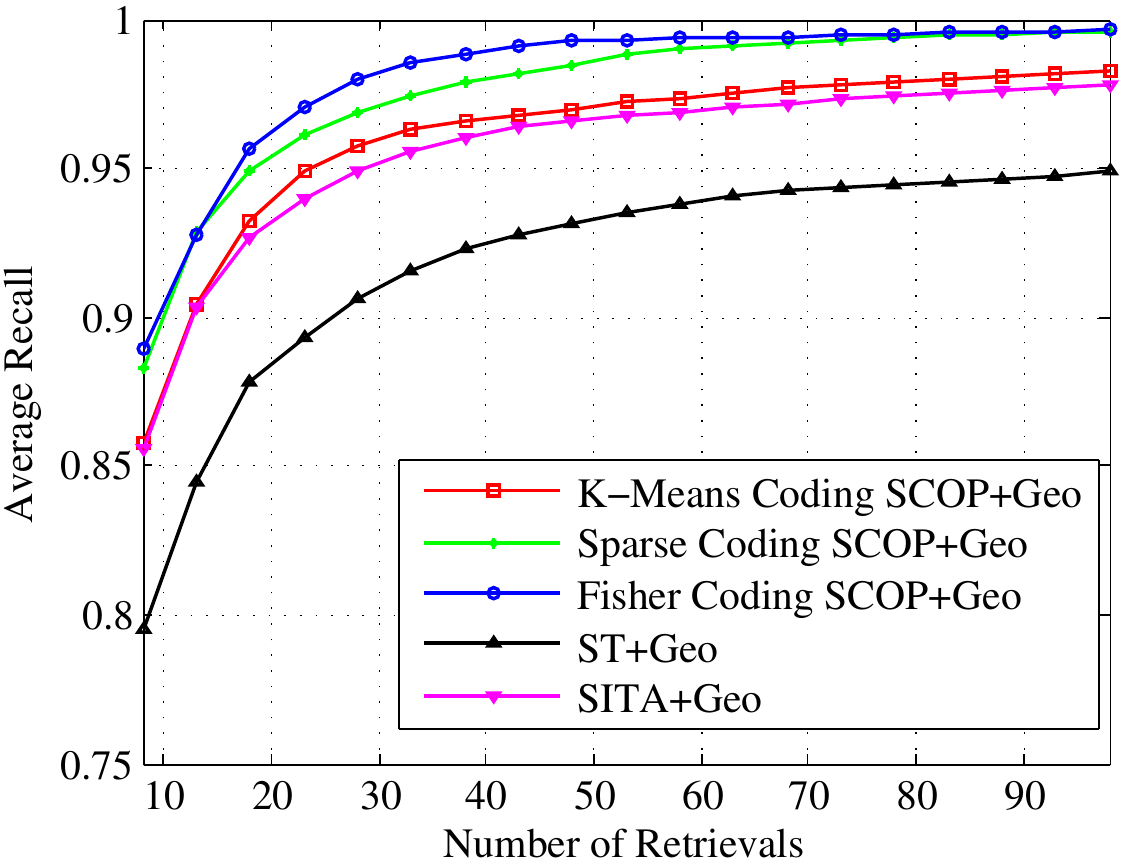}
  \caption{Average retrieval performances of different method on texture datasets. From left to right: UIUC, UMD, Brodatz datasets. The red, blue and green curves indicate our KM-SCOP+Geo, SC-SCOP+Geo, FC-SCOP+Geo, while the pink and dark curves indicate the result of SITA~\cite{sifre2013rotation} and Scatter Transform (ST)~\cite{xia2010shape}. The X-axis is the numbers of retrieval samples and the Y-axis is average recalls, which means the average of the retrieval true samples out of the true samples in the dataset.}
  \label{fig:ret_texture}
\end{figure*}

Fig.\ref{fig:ret_texture} shows the retrieval results of UIUC, UMD and Brodatz datasets in the left, middle and right respectively. The X-axis is the numbers of retrieval samples and the Y-axis is average recalls, which means the average of the retrieval true samples out of the true samples in the dataset. As we have observed that our algorithm works better when the training size is small, we can see that our results outperform others when we use one sample to retrieve the similar samples. More precisely, when the recall number is $39$ on the UIUC and UMD datasets, the retrieval rates of the proposed method with geodesics (KM-SCOP+Geo) are respectively $87.4\%$ and $88.3\%$, comparing with $61.2\%$ and $66.1\%$ of the scattering transform features with geodesics (ST+Geo) and $78.5\%$ and $87.0\%$ of the SITA approach with geodesics (SITA+Geo). When the recall number is $8$ on Brodatz dataset, our retrieval rate (FC-SCOP+Geo)is $88.9\%$ compared to $79.2\%$ and $85.3\%$ for ST+Geo and SITA+Geo.

It is interesting to notice that KM-SCOP+Geo performs better than SC-SCOP+Geo and FC-SCOP+Geo at UIUC and UMD datasets while FC-SCOP+Geo performs best at Brodatz dataset. The large difference between classes in UIUC and UMD datasets, caused by the scale variance and affine variance \emph{et.al.}, makes it difficult to retrieve by only one sample with a more complex model. However, the Brodatz dataset suffers no scale or affine variances, so we will not over fit this dataset with more complex model.

\begin{figure*}[htb!]
\centering
  \includegraphics[width=0.98\linewidth]{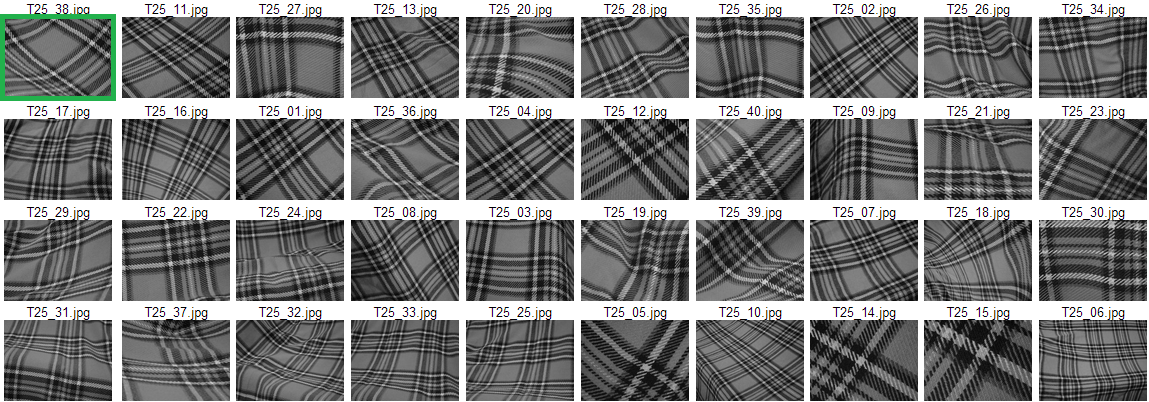}
  \caption{The retrieval result of one sample in Plaid in UIUC dataset}
\label{fig:uiucplaid}
\end{figure*}

The accurate retrieval result of one sample of Plaid in UIUC dataset is shown in Fig.\ref{fig:uiucplaid}. This sample is under 3D deformation, which is found to be a very tough task for retrieval, while our algorithm is totally correct to retrieve all of the similar samples in this dataset.

From the analysis of the above texture results, We conclude that our algorithm is effective to model the texture images. Due to the texture can be regarded as the arrangement of repetitive textons, including some deformation or translation, our algorithm models the textons by the geometric structures directly. Because the attributes of the shapes are invariant to most geometric deformation, our statistical histograms are robust to the deformation. More importantly, our method can not only depict both the small structures and large structures at the same time resorting to topographic map, but also describe the relationship of these structures. We also validate our method on the scene dataset which include the outdoor scene dataset and the remote sensing scene datasets in the next section.

\subsection{Experiments on scene description}

\begin{figure*}[htb!]
\centering
  \includegraphics[width=0.98\linewidth]{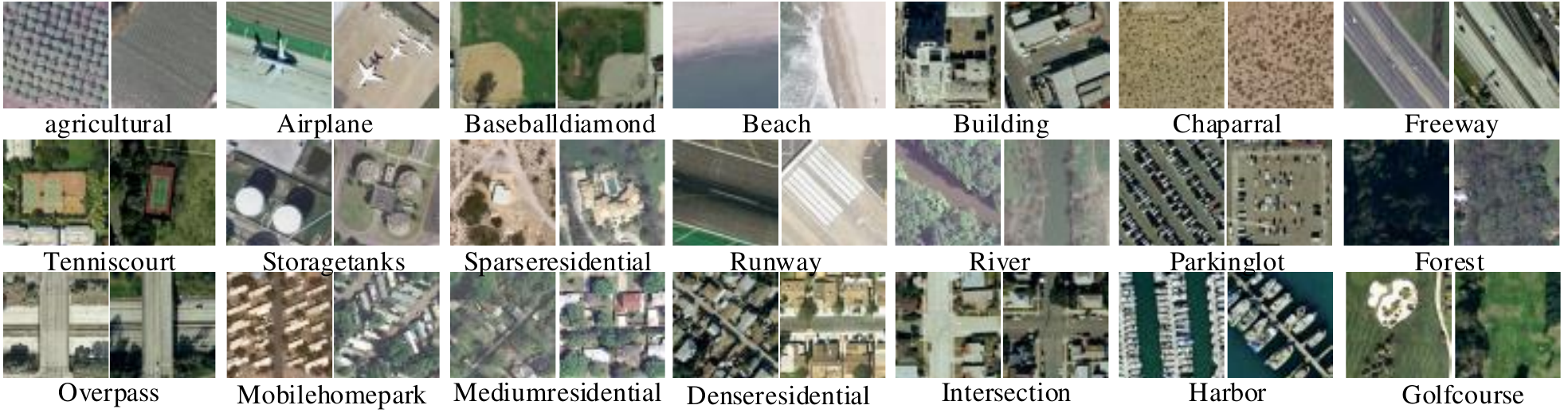}
  \caption{The illustration of UC Merced Lanuse dataset}
  \label{fig:remotes}
\end{figure*}

This section applies our algorithm into scene classification. The scene usually consists of significant objects and the surrounding environment. In our opinion, the surrounding environment could be regarded as the homogeneous texture while the objects could be regarded as kind of significant structures of the texture.
Due to this, we evaluate our method on three scene datasets, two of which are about remote sensing datasets and another one is about the outdoor scene dataset. The compared methods are tuned to use the suitable parameters. In our algorithm, parameters are set up according to information of the datasets, which are described below:
\begin{itemize}
  \item[-] \textbf{UC Merced Landuse dataset~\cite{yang2010bag}}: This dataset contains $21$ classes with $100$ images of size $256\times256$ per class. The images were manually extracted by the author from large images from the USGS National Map Urban Area Imagery collection for various urban areas around the country. The pixel resolution of this public domain imagery is $1$ foot. This dataset contains various classes of scenes, such as airplane, basement, city \emph{et al.}, which are shown in Fig.\ref{fig:remotes}. We can see that the scene are under different scales and rotations. The significant objects in the scene are arranged randomly, too. Besides, the parameter of interval $r$ between parent and children is $5$ by Eq.~\eqref{eq:areaper} and the number of cascaded ancestors $\tau/r$ is set to be $1$. We evaluate the method with $30$, $50$, $80$ samples per class for training, and the rest for testing.
  \item[-] \textbf{WHU-RS19~\cite{xia2010structural}}: This dataset contains $19$ classes with $50$ images of size $600\times600$ per class. This dataset includes images about airport, bridge, desert, forest, railway station, river \emph{et al.} Different from UC Merced Landuse dataset, the scene samples in this dataset are more complicated. For example, the airport contains crisscrossed runways, smooth lawn, plane stations and stopping planes. While the river samples contain winding river with trees around it. More details can be referred in \cite{xia2010structural}. The parameter of interval $r$ is $5$ by Eq.~\eqref{eq:areaper} and the number of cascaded ancestors $\tau/r$ is set to be $1$. We evaluate the method with $15$, $25$, $40$ samples per class for training, and the rest for testing.
  \item[-] \textbf{MIT Outdoor dataset~\cite{oliva2001modeling}}: This dataset contains $8$ different outdoor scenes, which are coast, forest, highway, insidecity, mountain, opencountry, street and tallbuilding, with over $300$ samples of size $256\times256$ per class. This kind of outdoor scene is very complex because the environment in the image is variable even within the same class. The significant objects are also suffer in variable view points or structures. Oliva~\cite{oliva2001modeling} \emph{et.al} has designed the specific feature "GIST" for this kind of dataset, which performs well and is used to compare with our algorithm. In our algorithm, the parameter of interval $r$ is $3$ by Eq.~\eqref{eq:areaper} and the number of cascaded ancestors $\tau/r$ is set to be $1$. We evaluate the method with $30\%$, $50\%$, $80\%$ samples per class for training, and the rest for testing.
\end{itemize}

The classification results of UC Merced Landuse Scene (UCMSce) dataset~\cite{yang2010bag}, WHU-RS19 dataset~\cite{xia2010structural} and MIT Outdoor Scene (MITOdSce) dataset~\cite{oliva2001modeling} are given in Tab.\ref{tab:scene}. In this table, the popular texture analysis methods of Bag-of-Visual-Words (BOVW)~\cite{Lazebnik2006}, SITA~\cite{xia2010shape} and Scattering transform (ST)~\cite{sifre2013rotation} are tested on all three datasets. The results of GIST~\cite{oliva2001modeling} for MIT Outdoor Scene dataset are also shown in this table.

We can see that our algorithm outperforms other methods for UCMSce dataset and WHU-RS19 dataset. The prominent classification results are achieved by our FC-SCOP when the training sizes are $30$, $50$ and $80$. For these two datasets, the SITA and ST do not perform as well as in the texture datasets. However, our method extracts the basic shape information from the scene images and constructs the SCOP features, which can not only characterize the basic elements of the scenes, but also describe the relationship of the basic elements, such as ``the bridge is on the river or the airplane is in the airport" by using the shape relationships. In our opinion, this relationships of the basic elements are essential for analyzing the scenes. Moreover, SCOP-SA outperforms SCOP-SS and SCOP-SAG, which verifies the opinion that statistical of more complicated hierarchical structures are not suitable for scene understanding here. So the number of cascaded ancestors is set to be $\tau/r=1$.

Another interesting finding is that, for MITOdSce dataset, the specific GIST~\cite{oliva2001modeling} defined by several properties for this kind of scenes, such as \emph{Degree of Naturalness, Degree of Openness, Degree of Roughness}, performs inferior to our method. Although these global properties are suitable to define an outdoor scene because the location of the objects keeps unchanged corresponding to the whole images, the local relationships between the structures are considered in our algorithm.


\begin{table*}[htb!]
\centering
\caption{Classification results with standard deviations on UCMSce~\cite{yang2010bag}, WHU-RS19~\cite{xia2010structural} and MITOdSce~\cite{oliva2001modeling}}
\label{tab:scene}
\tiny
\begin{tabular}{c|c|c|c|c|c|c|c|c|c}
  \hline
  & \multicolumn{3}{|c|}{UCMSce} & \multicolumn{3}{|c|}{WHU-RS19} &\multicolumn{3}{|c}{MITOdSce}\\\hline
  Training size                 & 30 & 50 & 80 & 15 & 25 & 40 & 30\% & 50\% & 80\% \\\hline
  BOVW~\cite{Lazebnik2006}      & 76.0$\pm$1.2 & 81.5$\pm$1.1 & 85.4$\pm$1.6 &77.5$\pm$1.7 &82.7$\pm$1.5 & 86.2$\pm$2.4 & 83.7$\pm$0.6 & 84.9$\pm$0.8 & 85.8$\pm$1.3\\
  GIST~\cite{oliva2001modeling} &-   & - &-  &-   & - &- &  83.0$\pm$0.7    &85.2$\pm$0.8 & 87.0$\pm$1.4 \\
  Yang~\cite{yang2010bag}       &-   & -     & 81.2 &-   & - &-  &-   & - &- \\
  SITA~\cite{xia2011texture}    &71.6$\pm$ 1.3 & 78.0$\pm$1.2 & 83.52$\pm$1.0 &70.3$\pm$1.9 & 76.7$\pm$1.8 & 80.42$\pm$1.8 & 78.6$\pm$0.8 & 80.5$\pm$0.9 & 82.0$\pm$1.4\\
  ST~\cite{sifre2013rotation}   & 77.7$\pm$1.2 & 80.7$\pm$1.1 & 82.6$\pm$1.7 &66.3$\pm$2.0 & 71.4$\pm$1.9 & 74.2$\pm$3.1 & 60.3$\pm$1.3 & 61.9$\pm$1.1 &63.1$\pm$1.9 \\\hline
  SCOP-SS &72.3$\pm$1.1 & 78.5$\pm$1.2 & 83.1$\pm$ 1.7
            &71.6$\pm$1.6 & 77.6$\pm$1.8 & 82.5$\pm$ 2.3
            &70.3$\pm$0.9 & 73.5$\pm$1.1 & 76.2$\pm$ 1.5\\
  SCOP-SA &76.1$\pm$1.1 & 80.8$\pm$1.1 & 84.2$\pm$ 1.7
            &76.5$\pm$1.2 & 76.4$\pm$6.7 & 80.6$\pm$ 4.2
            &71.9$\pm$0.9 & 74.5$\pm$0.9 & 76.4$\pm$ 1.8  \\
  SCOP-SAG &75.9$\pm$1.0 & 80.2$\pm$1.1 & 83.3$\pm$ 1.7
            &77.4$\pm$0.2 & 78.9$\pm$5.9 & 79.9$\pm$ 4.8
            &70.2$\pm$0.9 & 72.6$\pm$1.0 & 74.8$\pm$ 1.6\\
  SCOP-SAS &69.2$\pm$1.3 & 73.7$\pm$1.2 & 77.7$\pm$ 1.8
            &71.7$\pm$0.5 & 71.8$\pm$6.6 & 74.8$\pm$ 3.9
            &66.3$\pm$0.9 & 69.3$\pm$1.0 & 71.7$\pm$ 1.8\\
  KM-SCOP  &80.1$\pm$1.0 & 84.5$\pm$1.0 & 88.5$\pm$ 1.4
            &82.5$\pm$1.4 & 86.3$\pm$1.6 & 89.2$\pm$ 2.1
            &75.0$\pm$0.9 & 78.0$\pm$1.0 & 80.1$\pm$ 1.6\\ \hdashline
  SC-SCOP & 81.7$\pm$1.1 & 86.2$\pm$1.0 & 89.7$\pm$ 1.4
                & 84.3$\pm$1.4 & 87.0$\pm$1.4 & 91.5$\pm$ 2.0
                & 81.9$\pm$0.8 & 84.0$\pm$0.8 & 85.9$\pm$1.4\\ \hdashline
  FC-SCOP & \textbf{85.0$\pm$1.0}  & \textbf{89.2$\pm$0.8} & \textbf{92.0$\pm$1.4}
                & \textbf{87.1$\pm$1.4}  & \textbf{90.9$\pm$1.2} & \textbf{93.3$\pm$1.7}
	           & \textbf{85.3$\pm$0.7}  & \textbf{86.8$\pm$0.7} & \textbf{88.0$\pm$1.3} \\ \hdashline

  FC-SCOP+MS & \textbf{86.9$\pm$1.0}  & \textbf{90.6$\pm$0.9} & \textbf{93.1$\pm$1.2}
    & \textbf{88.1$\pm$1.4}  & \textbf{92.0$\pm$1.2} & \textbf{94.7$\pm$1.5}
    & \textbf{ 85.5+$\pm$0.8} & \textbf{87.5$\pm$0.7}  & \textbf{88.4$\pm$1.2} \\
  \hline
\end{tabular}
\end{table*}

We can conclude this as, in the remote sensing scene images, the basic elements, such as boats, cars, trees, \emph{et al.} are projected into the bag of shapes by their SCOP features. It is more general to model the basic elements and their relationship based on the topographical map, which is robust and discriminated to scene classification.

\section{Conclusion}
\label{SEC:con}
This paper proposed a texture analysis method by investigating the shape co-occurrence patterns with several coding strategy. According to the fundamental analysis of shape-based image representation\cite{xia2010shape}, our algorithm is robust to most of the geometric deformation and illumination changes. More importantly, our algorithm constructs the relationship of local structure by shape co-occurrence patterns, then counts the statistics of the patterns. The comparisons of retrieval and classification results show the high performance of our algorithm. Remark that the SITA\cite{xia2010shape} is an special case of our method except the marginal distributions of individual attributes of shapes are considered.

Note that our algorithm performs well on both the texture dataset and the scene dataset. Especially on the texture datasets, the SCOPs model the complex texture structures well and improve a lot from using the high order information. Besides, in scene datasets, thousands of objects appear and constitute the scenes, which is known for hardly understanding. While our method depicts the geometrical aspects of textons or elements as well as being
robust to geometric and illumination changes.

Due to the robust description of the local structures and their co-occurrence patterns, the further investigation will be concentrated on the more complex texture recognition and large scene image interpretation or segmentation. Moreover, a more effective combination of patterns by using deep learning is under research.

\section*{Acknowledgment}
This research is supported by the National Natural Science Foundation of China under the contracts No.91338113 and No.41501462.

\bibliographystyle{ieeetr}
\bibliography{egbib,refer_icpr}

\begin{thebibliography}{10}

\bibitem{Jeanponce02computervision}
D.~A. Forsyth and J.~Ponce, {\em Computer Vision: A Modern Approach}.
\newblock {Prentice Hall}, 2002.

\bibitem{texture_Haralick73}
R.~M. Haralick, K.~Shanmugam, and I.~Dinstein, ``Textural features for image
  classification,'' {\em IEEE Transaction on Systems, Man and Cybernetics},
  vol.~SMC-3, no.~6, pp.~610--621, 1973.

\bibitem{LeungM01}
T.~K. Leung and J.~Malik, ``Representing and recognizing the visual appearance
  of materials using three-dimensional textons,'' {\em International Journal on
  Computer Vision}, vol.~43, no.~1, pp.~29--44, 2001.

\bibitem{Portilla2000}
J.~Portilla and E.~P. Simoncelli, ``A parametric texture model based on joint
  statistics of complex wavelet coefficients,'' {\em International Journal on
  Computer Vision}, vol.~40, no.~1, pp.~49--70, 2000.

\bibitem{texture_Lazebnik2}
S.~Lazebnik, C.~Schmid, and J.~Ponce, ``A sparse texture representation using
  local affine regions,'' {\em IEEE Transactions on Pattern Analysis and
  Machine Intelligence}, vol.~27, no.~8, pp.~1265--1278, 2005.

\bibitem{xia2010shape}
G.-S. Xia, J.~Delon, and Y.~Gousseau, ``Shape-based invariant texture
  indexing,'' {\em International Journal of Computer Vision}, vol.~88, no.~3,
  pp.~382--403, 2010.

\bibitem{Pedersen2013}
K.~Steenstrup~Pedersen, K.~Stensbo-Smidt, A.~Zirm, and C.~Igel, ``Shape index
  descriptors applied to texture-based galaxy analysis,'' in {\em Proceedings
  of the IEEE International Conference on Computer Vision}, pp.~2440--2447,
  2013.

\bibitem{Li2013}
R.~Li and E.~H. Adelson, ``Sensing and recognizing surface textures using a
  gelsight sensor,'' in {\em Proceedings of the IEEE Conference on Computer
  Vision and Pattern Recognition}, pp.~1241--1247, 2013.

\bibitem{Matthews2013}
T.~Matthews, M.~S. Nixon, and M.~Niranjan, ``Enriching texture analysis with
  semantic data,'' in {\em Proceedings of the IEEE Conference on Computer
  Vision and Pattern Recognition}, pp.~1248--1255, 2013.

\bibitem{sifre2013rotation}
L.~Sifre and S.~Mallat, ``Rotation, scaling and deformation invariant
  scattering for texture discrimination,'' in {\em Proceedings of the IEEE
  Conference on Computer Vision and Pattern Recognition}, pp.~1233--1240, 2013.

\bibitem{liu2011sorted}
L.~Liu, P.~Fieguth, G.~Kuang, and H.~Zha, ``Sorted random projections for
  robust texture classification,'' in {\em IEEE International Conference on
  Computer Vision}, pp.~391--398, 2011.

\bibitem{liu2012}
L.~Liu and P.~Fieguth, ``{Texture classification from random features},'' {\em
  IEEE Transactions on Pattern Analysis and Machine Intelligence}, vol.~34,
  no.~3, pp.~574--586, 2012.

\bibitem{xu2010new}
Y.~Xu, X.~Yang, H.~Ling, and H.~Ji, ``A new texture descriptor using
  multifractal analysis in multi-orientation wavelet pyramid,'' in {\em IEEE
  Conference on Computer Vision and Pattern Recognition}, pp.~161--168, 2010.

\bibitem{Julesz}
B.~Julesz, ``Textons, the elements of texture perception, and their
  interactions.,'' {\em Nature}, vol.~290, no.~5802, pp.~91--97, 1981.

\bibitem{Zhu2005textons}
S.-C. Zhu, C.-E. Guo, Y.~Wang, and Z.~Xu, ``What are textons?,'' {\em
  International Journal of Computer Vision}, vol.~62, no.~1-2, pp.~121--143,
  2005.

\bibitem{Peyre08}
G.~Peyr\'e, ``Texture synthesis with grouplets,'' {\em IEEE Transactions on
  Pattern Analysis and Machine Intelligence}, vol.~32, no.~4, pp.~733--746,
  2009.

\bibitem{Varma2005}
M.~Varma and A.~Zisserman, ``A statistical approach to texture classification
  from single images,'' {\em International Journal on Computer Vision},
  vol.~62, no.~1-2, pp.~61--81, 2005.

\bibitem{Ojala2002}
T.~Ojala, M.~Pietikainen, and T.~Maenpaa, ``Multiresolution gray-scale and
  rotation invariant texture classification with local binary patterns,'' {\em
  IEEE Transactions on Pattern Analysis and Machine Intelligence}, vol.~24,
  no.~7, pp.~971--987, 2002.

\bibitem{xu2009viewpoint}
Y.~Xu, H.~Ji, and C.~Ferm{\"u}ller, ``Viewpoint invariant texture description
  using fractal analysis,'' {\em International Journal of Computer Vision},
  vol.~83, no.~1, pp.~85--100, 2009.

\bibitem{ji2013wavelet}
H.~Ji, X.~Yang, H.~Ling, and Y.~Xu, ``Wavelet domain multi-fractal analysis for
  static and dynamic texture classification,'' {\em IEEE Transactions on Image
  Processing}, vol.~22, no.~1, pp.~286--299, 2013.

\bibitem{nguyen2011visual}
H.-G. Nguyen, R.~Fablet, and J.-M. Boucher, ``Visual textures as realizations
  of multivariate log-gaussian cox processes,'' in {\em IEEE Conference on
  Computer Vision and Pattern Recognition}, pp.~2945--2952, 2011.

\bibitem{qi2014Pairwise}
X.~Qi, R.~Xiao, C.-g. Li, Y.~Qiao, J.~Guo, and X.~Tang, ``{Pairwise rotation
  iInvariant co-occurrence local binary pattern},'' {\em IEEE Transactions on
  Pattern Analysis and Machine Intelligence}, vol.~36, pp.~2199--2213, nov
  2014.

\bibitem{Cimpoi2015}
M.~Cimpoi, S.~Maji, and A.~Vedaldi, ``{Deep filter banks for texture
  recognition and segmentation},'' {\em IEEE Conference on Computer Vision and
  Pattern Recognition}, pp.~3828--3836, nov 2015.

\bibitem{Lafarge09}
F.~Lafarge, G.~Gimel'farb, and X.~Descombes, ``Geometric feature extraction by
  a multimarked point process,'' {\em IEEE Transactions on Pattern Analysis and
  Machine Intelligence}, vol.~32, no.~9, pp.~1597--1609, 2010.

\bibitem{lazebnik2005sparse}
S.~Lazebnik, C.~Schmid, and J.~Ponce, ``A sparse texture representation using
  local affine regions,'' {\em IEEE Transactions on Pattern Analysis and
  Machine Intelligence}, vol.~27, no.~8, pp.~1265--1278, 2005.

\bibitem{texture_Larry}
L.~Davis, ``Polarograms: a new tool for image texture analysis,'' {\em Pattern
  Recognition}, vol.~13, no.~3, pp.~219--223, 1981.

\bibitem{pietikainen2000rotation}
M.~Pietik{\"a}inen, T.~Ojala, and Z.~Xu, ``Rotation-invariant texture
  classification using feature distributions,'' {\em Pattern Recognition},
  vol.~33, no.~1, pp.~43--52, 2000.

\bibitem{Cimpoi2014}
M.~Cimpoi, S.~Maji, I.~Kokkinos, S.~Mohamed, and A.~Vedaldi, ``{Describing
  textures in the wild},'' {\em IEEE Conference on Computer Vision and Pattern
  Recognition}, pp.~3606--3613, nov 2014.

\bibitem{Simonyan14c}
K.~Simonyan and A.~Zisserman, ``Very deep convolutional networks for
  large-scale image recognition,'' {\em Computing Research Repository},
  vol.~abs/1409.1556, 2014.

\bibitem{monasse2000fast}
P.~Monasse and F.~Guichard, ``Fast computation of a contrast-invariant image
  representation,'' {\em IEEE Transactions on Image Processing}, vol.~9, no.~5,
  pp.~860--872, 2000.

\bibitem{xia2014texture}
G.~Liu, G.-S. Xia, W.~Yang, and L.~Zhang, ``Texture analysis by using shapes
  co-occurrence patterns,'' in {\em International Conference on Pattern
  Recognition}, pp.~1--6, 2014.

\bibitem{caselles1999topographic}
V.~Caselles, B.~Coll, and J.-M. Morel, ``Topographic maps and local contrast
  changes in natural images,'' {\em International Journal of Computer Vision},
  vol.~33, no.~1, pp.~5--27, 1999.

\bibitem{Monasse08}
V.~Caselles and P.~Monasse, {\em Geometric Description of Topographic Maps and
  Applications to Image Processing}.
\newblock Lecture Notes in Mathematics, Springer, 2009.

\bibitem{ren2013histograms}
X.~Ren and D.~Ramanan, ``Histograms of sparse codes for object detection,'' in
  {\em IEEE Conference on Computer Vision and Pattern Recognition},
  pp.~3246--3253, 2013.

\bibitem{mairal2009online}
J.~Mairal, F.~Bach, J.~Ponce, and G.~Sapiro, ``Online dictionary learning for
  sparse coding,'' in {\em International Conference on Machine Learning},
  pp.~689--696, 2009.

\bibitem{mairal2014sparse}
J.~Mairal, F.~Bach, and J.~Ponce, ``Sparse modeling for image and vision
  processing,'' {\em arXiv preprint arXiv:1411.3230}, 2014.

\bibitem{bailey1994fitting}
T.~L. Bailey, C.~Elkan, {\em et~al.}, ``Fitting a mixture model by expectation
  maximization to discover motifs in bipolymers,'' 1994.

\bibitem{JEGOU-2011-633013}
H.~J{\'e}gou, F.~Perronnin, M.~Douze, J.~S{\'a}nchez, P.~P{\'e}rez, and
  C.~Schmid, ``{Aggregating local image descriptors into compact codes},'' {\em
  IEEE Transactions on Pattern Analysis and Machine Intelligence}, 2011.

\bibitem{chang2011libsvm}
C.-C. Chang and C.-J. Lin, ``Libsvm: a library for support vector machines,''
  {\em ACM Transactions on Intelligent Systems and Technology}, vol.~2, no.~3,
  p.~27, 2011.

\bibitem{brodatz1966textures}
P.~Brodatz, {\em Textures: a photographic album for artists and designers},
  vol.~66.
\newblock Dover New York, 1966.

\bibitem{crosier2008texture}
M.~Crosier and L.~D. Griffin, ``Texture classification with a dictionary of
  basic image features,'' in {\em IEEE Conference on Computer Vision and
  Pattern Recognition}, pp.~1--7, 2008.

\bibitem{xia2011texture}
G.-S. Xia, F.~Yuan, {\em et~al.}, ``Texture segmentation by grouping ellipse
  ensembles via active contours,'' in {\em British Machine Vision Conference},
  pp.~1--11, 2011.

\bibitem{oliva2001modeling}
A.~Oliva and A.~Torralba, ``Modeling the shape of the scene: A holistic
  representation of the spatial envelope,'' {\em International Journal of
  Computer Vision}, vol.~42, no.~3, pp.~145--175, 2001.

\bibitem{yang2010bag}
Y.~Yang and S.~Newsam, ``Bag-of-visual-words and spatial extensions for
  land-use classification,'' in {\em Proceedings of the 18th SIGSPATIAL
  International Conference on Advances in Geographic Information Systems},
  pp.~270--279, 2010.

\bibitem{xia2010structural}
G.-S. Xia, W.~Yang, J.~Delon, Y.~Gousseau, H.~Sun, and H.~Ma{\^\i}tre,
  ``Structural high-resolution satellite image indexing,'' in {\em ISPRS TC VII
  Symposium-100 Years}, vol.~38, pp.~298--303, 2010.

\bibitem{Lazebnik2006}
S.~Lazebnik, C.~Schmid, and J.~Ponce, ``Beyond bags of features: Spatial
  pyramid matching for recognizing natural scene categories,'' in {\em IEEE
  Conference on Computer Vision and Pattern Recognition}, vol.~2,
  pp.~2169--2178, 2006.

\end{thebibliography}

\end{document}